\documentclass[final]{cvpr}
\usepackage{times}
\usepackage{epsfig}
\usepackage{graphicx}
\usepackage{amsmath}
\usepackage{amssymb}
\usepackage{multirow}
\usepackage[page,toc,titletoc,title]{appendix}
\usepackage{titletoc}
\usepackage{booktabs}
\usepackage{bm}
\usepackage{dsfont}
\usepackage{subfigure}
\usepackage{pifont}
\usepackage{parskip}
\usepackage[nocompress]{cite}
\usepackage[pagebackref=true,breaklinks=true,colorlinks,bookmarks=false]{hyperref}
\usepackage{cleveref}
\usepackage{textcomp}
\usepackage[numbers,sort]{natbib} % sort inline citations
\usepackage{pifont}% http://ctan.org/pkg/pifont
\usepackage[table,xcdraw]{xcolor}
\usepackage{longtable}
\usepackage{pdflscape}
\usepackage{caption}
\usepackage{wrapfig}

\usepackage[pagebackref=true,breaklinks=true,colorlinks,bookmarks=false]{hyperref}

\def\cvprPaperID{2958} % *** Enter the CVPR Paper ID here
\def\confYear{CVPR 2021}

\newcommand{\cmark}{\ding{51}}%
\newcommand{\xmark}{\ding{55}}%

\newcommand{\TQN}{{\tt TQN}\xspace}
\newcommand{\SUFB}{stochastically updated feature bank\xspace}
\newcommand{\sufb}{feature bank\xspace}

\newcommand{\bq}{${\bf q}$\xspace}

\newcommand{\br}{${\bf r}$\xspace}

\renewcommand{\paragraph}[1]{\noindent{\bf #1}\quad}

\newif\ifarxiv
\newcommand{\sref}[1]{%
	\ifarxiv%
	{\Cref{#1}}%
	\else%
		{the extended version~\cite{zhang21}}%
	\fi%
}
\arxivtrue  % comment out for arXiv
\begin{document}
	
	\title{Temporal Query Networks for Fine-grained Video Understanding}
	
	\author{Chuhan Zhang\\
		University of Oxford\\
		{\tt\small czhang@robots.ox.ac.uk}	
		\and
		Ankush Gupta\\
		DeepMind, London\\
		{\tt\small ankushgupta@google.com}
		\and
		Andrew Zisserman\\
		University of Oxford\\
		{\tt\small az@robots.ox.ac.uk}}
	
	\maketitle

	\urlstyle{same}

	\begin{abstract}
		Our objective in this work is fine-grained classification of actions in 
		untrimmed videos, where the actions may be temporally extended or may span only 
		a few frames of the video. We cast this into a query-response mechanism, where 
		each query addresses a particular question, and has its own response label set. 
		
		We make the following four contributions: (i)~We propose a new model---a Temporal Query
		Network---which enables the query-response functionality, and a structural 
		understanding of fine-grained actions. It attends to relevant segments for 
		each query with a temporal attention mechanism, and can be trained using only 
		the labels for each query.
		(ii)  We propose a new way---stochastic feature bank update---to train a network 
		on videos of various lengths with the dense sampling required to respond to 
		fine-grained queries.
		(iii)~We compare the TQN to other architectures and text supervision methods, 
		and analyze their pros and cons. Finally, (iv)~we evaluate the method
		extensively on the FineGym and Diving48 benchmarks for fine-grained action classification and %significantly 
		surpass the state-of-the-art using only RGB features.\\\\ 
		Webpage: %{\small\url{http://www.robots.ox.ac.uk/~vgg/research/tqn/}}.
		\href{http://www.robots.ox.ac.uk/~vgg/research/tqn/}{\nolinkurl{http://www.robots.ox.ac.uk/~vgg/research/tqn/}}.

		% Our objective is to action recognition in videos, in particular for fine-grained action which needs temporal reasoning. While there has been great progress in recognizing action using 3D convolutional network, processing dense and fine-grained information is still a challenging task. In this work, we make the following contributions: i) We investigate how important dense sampling is for rapid movement and propose a way to train a network on videos of various lengths with dense sampling. ii) We propose a new architecture -- Temporal Query Network which utilizes a small amount of text descriptions to build structural understanding of fine-grained action. It attends to relevant segments with a temporal attention mechanism hence achieves excellent recognition accuracy. iii) We compare different architectures which learn to model action sequence through text sequence and analyze their pros and cons, iv) We evaluate our method extensively on FineGym, Diving48 and SomethingSomething and demonstrate state-of-the-art by only using RGB features.
	\end{abstract}

	\begin{figure}
		\vspace{4mm}
		% \includegraphics[width=\linewidth]{figs/TQN-teaser.pdf}
		% \caption{{\bf Fine-grained action recognition.} Object and background cues from only a few frames can inform classic action recognition (\eg, in the Kinetics dataset~\cite{kinetics}) where visually distinct activities are to be distinguished ({\em\bf left}). However, such cues are not effective for fine-grained action recognition (\eg, in the FineGym dataset~\cite{shao2020finegym}) ({\em\bf right}), where subtle differences and details matter---{\em\bf[right, top two]:} relative location at start and end of the sequence on the beam; {\em\bf [right, bottom two]:} direction of facing and the number of twists. Further, the classes are multi-part labels with shared sub-structure instead of a single activity name, \eg, {\em ``split leap''} vs. {\em ``split jump''} and {\em ``salto backward stretch with 1.5 twists''} vs. {\em ``salto forward stretch with 1.0 twist.''} Finally, sub-sampling frames (as with five frames above) causes temporal aliasing which loses details useful for accurate recognition, necessitating dense input sampling over the entire video sequence. We develop a novel video model to address these challenges associated with fine-grained video recognition.}%
		\includegraphics[width=\linewidth]{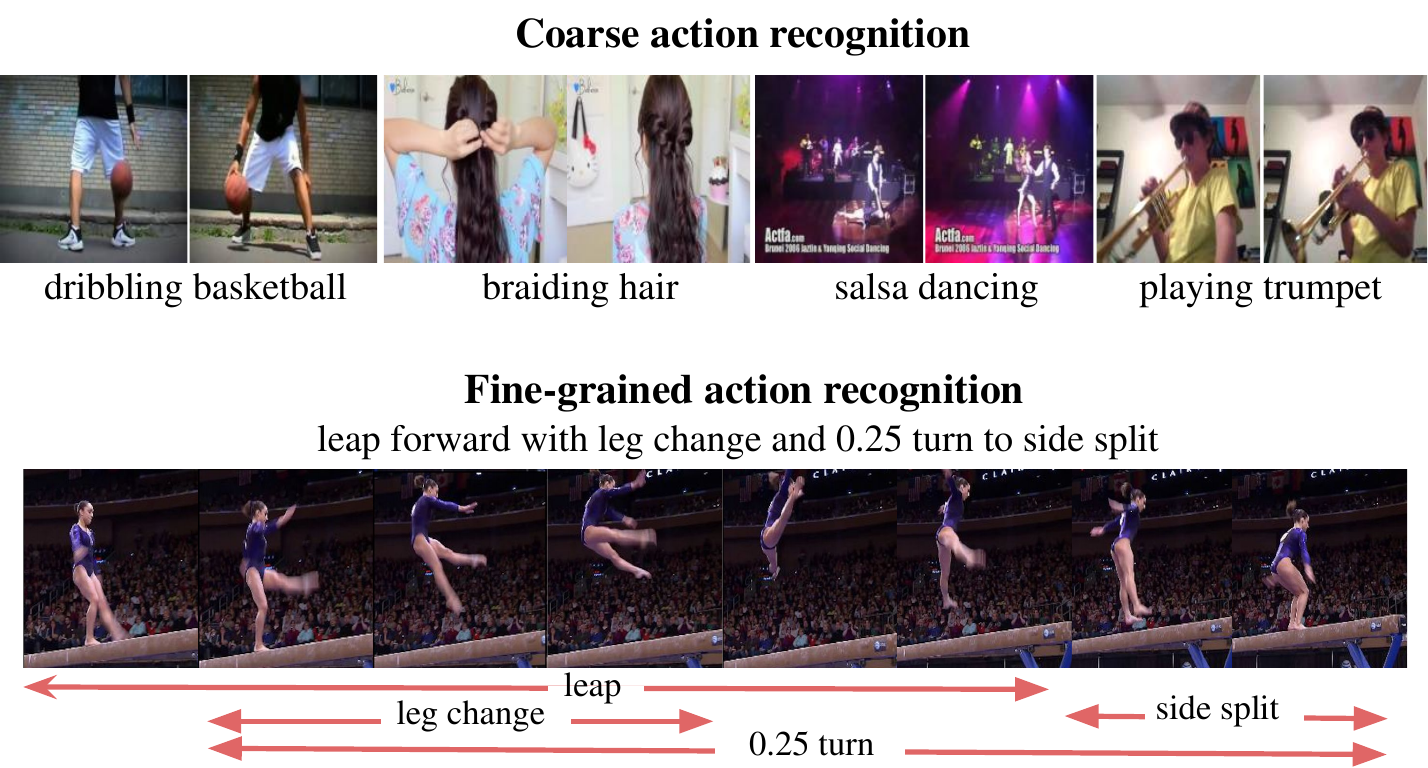}
		\caption{{\bf Coarse vs.\ fine-grained action recognition.} 
			{\em\bf Top:} Object and background cues from only a few frames can inform classic
			coarse-grained action recognition in datasets like Kinetics~\cite{kinetics}, where visually distinct activities are to be distinguished. 
			{\em\bf Bottom:} However, for finer-grain classification which depends on subtle differences in pose, the specific sequence, duration and 
			number of certain sub-actions, as for the gymnastics sequence above, requires reasoning about events at varying temporal scales and attention to fine details.
			We develop a novel query-based video network and a training framework for such fine-grained temporal reasoning.}%
		%	\caption{{\bf Coarse vs.\ fine-grained action recognition.} 
		%{\em\bf Left:} In datasets like Kinetics dataset~\cite{kinetics}, 
		%object and background cues from only a few frames can inform classic
		%coarse-grained action recognition where visually distinct activities are to be distinguished. 
		%{\em\bf Right:} Such cues are not effective for fine-grained action
		%recognition (\eg, in the FineGym dataset~\cite{shao2020finegym})
		%where subtle differences and details matter, \eg, in determining the 
		%direction of facing or number of twists, and require dense attention to all frames of the video ** the rest of this is (1) too long, (2) too detailed, and (3) unwise -- is it showing sub-sampling so that the action can not be seen? That is not a good idea, as we can't see what the task is ** Further, the classes are
		%multi-part labels with shared sub-structure instead of a single
		%activity name, \eg, {\em ``salto backward stretch with 1.5 twists''}
		%and {\em ``salto forward stretch with 1.0 twist.''} Finally,
		%sub-sampling frames (as with five frames above) causes temporal
		%aliasing which loses details useful for accurate recognition,
		%necessitating dense input sampling over the entire video sequence. We
		%develop a novel video model to address these challenges associated
		%with fine-grained video recognition.}%
		\label{f:teaser}
	\end{figure}
	
	\section{Introduction}\label{sec:intro}
	Imagine that you wish to answer particular questions about a video. These questions could be quite general, \eg,
	``what instrument is being played?'', quite specific, \eg, ``do people shake hands?'', or require a composite answer,
	\eg, ``how many somersaults, if any,  are performed in this video, and where?''. 
	Answering these questions will in general require attending to the entire video (to ensure that nothing is missed), 
	and the response is {\em query dependent}.
	Further, the response may depend on only a very few frames where a subtle action occurs.
	With such video understanding capability, it is possible to effortlessly %analyze vast video collections, or 
	carry out regular video metrology such as performance evaluation in sports training, or issuing reports on video logs.

	The objective of this paper is a network and training framework that
	will enable questions of various granularity to be answered on a
	video. Specifically, we consider {\em untrimmed videos} and train with {\em weak
		supervision}, meaning that at training time we are not provided with
	the temporal localization information for the response. To this end, we introduce a new Transformer-based~\cite{vaswani2017attention}
	video network architecture, the {\em Temporal Query Network} (\TQN), for fine-grained action classification. % question and answering.
	The \TQN ingests a video and a pre-defined set of {\em queries} and outputs {\em responses} for each query, where the response is query dependent.
	
	The queries act as `experts' that are able to pick out from the video the temporal segments required for their response. Since the temporal position of the response is unknown, they must examine the entire duration of the video and be able
	to ignore irrelevant content, in a similar manner to a `matched filter'~\cite{turin1960introduction}. Furthermore, since the duration of response
	segments may only be a few frames, excessive temporal aggregation  (for example, by average pooling the entire untrimmed video) may lose the signal in the noise.
	
	As the \TQN must attend {\em densely} to the video frames for answering specific queries, and cannot sub-sample in time, we also introduce
	a {\em stochastically updated feature bank} so that the model can be trained beyond the constraints imposed by finite
	GPU memory. For this we use a temporal feature bank in which features from densely sampled contiguous temporal segments are cached over the course of training, and only a random subset of these features is computed online and backpropagated through in each training iteration.
	
	We demonstrate the \TQN on two fine-grained action recognition datasets with untrimmed video sequences:
	FineGym~\cite{shao2020finegym} and Diving48~\cite{li2018resound}. Both of these datasets share
	the following challenges: (i) object and backgrounds cannot be used to inform classification, as is possible for more coarse-grained action recognition datasets, \eg, Kinetics~\cite{kinetics} and UCF-101~\cite{ucf101} (see \Cref{f:teaser}).
	(ii) subtle differences in actions, relative spatial orientations and temporal ordering of objects/actors need
	to be distinguished. (iii) events have a short duration of approx.\ 0.3 seconds in video clips which are typically 6-10 seconds long, and can be as much as 30 seconds in length. (iv) Finally,
	the duration and position of events vary and is unknown in training. This lack
	of alignment between text-description (labels) and videos means that that supervision is weak. 
	
	\paragraph{Summary of contributions:} (i)~we introduce a new model---a Temporal 
	Query Network (\TQN)--which enables query-response functionality on untrimmed 
	videos. It can be trained using only the labels for each query. We show how 
	fine-grained video classification can be cast as a query-response task.
	(ii)~We propose a new way---stochastic feature bank update---to train a network 
	on videos of various lengths with the dense sampling required to respond to 
	fine-grained queries.
	(iii)~We compare the \TQN to other architectures and text supervision methods, 
	and analyze their pros and cons. 
	Finally, (iv)~we evaluate the method extensively on the 
	FineGym~\cite{shao2020finegym} and Diving48~\cite{li2018resound} benchmarks for fine-grained action classification. 
	We demonstrate the benefits of the \TQN and stochastic feature bank update over 
	baselines and with ablations, and the importance of extended and dense temporal context. The \TQN 
	with stochastic feature bank update training surpass the state-of-the-art on 
	these two benchmarks using only RGB features.

	\section{Related Work}\label{s:rel_work}
	\paragraph{Action Recognition.}
	Convolutional neural network have been widely used in action recognition recently, including both 2D networks like the two-stream~\cite{two-stream}, TSN~\cite{tsn}, TRN~\cite{trn}, TSM~\cite{tsm}, TPN~\cite{tpn}, and 3D networks like LTC~\cite{ltc}, I3D~\cite{i3d}, S3D~\cite{s3d}, SlowFast~\cite{slowfast}, X3D~\cite{x3d}.  Progress in architectures has led to a steadily improved performance on both coarse and fine-grained action datasets~\cite{kinetics,ucf101,breakfast,epic-kitchen}. Despite this success, challenges remain:  fine-grained action recognition without objects and background biases~\cite{danceinmall,li2018resound},  long-term action understanding~\cite{featbank,videolstm}, and distinguishing actions with subtle differences~\cite{shao2020finegym}.

	\paragraph{Long-Term Video Understanding.}
	Early work used RNNs like LSTM~\cite{lstm} for context-modeling in long videos~\cite{Li18,li2019beyond}. More recently, the Transformer~\cite{vaswani2017attention} architecture has been widely adopted for vision tasks due to its advantage in modeling long-term dependencies. The combination of ConvNets and Transformer is applied not only for images~\cite{detr,virtex,ViT,ImageGPT,actbert}, but also on video tasks  including representation learning~\cite{videobert,mmt,CBT}, and action classification~\cite{nonlocal,actiontfm,featbank}.

	\paragraph{Video-Text Representation Learning.}
	Videos are naturally rich in modalities, and text extracted from associated captions, audio, and transcripts is often used for video representation learning.  ~\cite{huang2016ctc,bojanowski2015weakly} use text as weak supervision to localize actions through alignment, but require text to have the same order as actions. \cite{arnab2020uncertainty,hahn2018learning} learn to localize and detect action from sparse text labels, while~\cite{gao2017tall} focuses on localizing actions in untrimmed videos by aligning free-form sentences, whereas we learn to answer specific questions with a pre-defined response set.
	Text is also used in self-supervised text-video representation learning~\cite{howto100m,milnce,videobert}, or for supervised tasks like video retrieval~\cite{mmt,Wray_2019_ICCV,Chen_2020_CVPR,Liu2019a}.
	%
	%MMT~\cite{gabeur2020mmt}, and \cite{Wray_2019_ICCV}.
	%Other text-based video retrieval methods: \cite{Liu2019a,Chen_2020_CVPR,miech2019howto100m,miech19endtoend}
	%and other works from: \url{https://www.robots.ox.ac.uk/~vgg/challenges/video-pentathlon/}
	
	\paragraph{Overcoming Memory Constraints in Frame Sampling.}
	A common way to extract features from a video is by sampling a fixed number of frames, usually less than 64~\cite{s3d,i3d,tsm}. However, such coarse sampling of frames is not sufficient, especially for fine-grained actions in untrimmed videos~\cite{ava,charades,li2018resound,shao2020finegym}.
	One common solution is to use pre-trained features\cite{featbank,li2017temporal,miech2017learnable,tang2018non,garcia2020knowledge}, but this relies on good initializations and ensuring a small domain gap. While another solution focuses on extracting key frames from untrimmed videos~\cite{SCsampler,gong2014diverse}.
	
	%\paragraph{Combining ConvNets and Attention Mechanisms.}
	%Early work used attention to bridge between modalities~\cite{Li18}. More recently, 
	%the attention mechanism provided by transformers~\cite{vaswani2017attention} has become increasingly popularity in vision tasks due to its advantage of modelling long-term dependencies. The combination of ConvNets and an
	%attention module has been used for image classification~\cite{detr,virtex,ViT,ImageGPT,actbert}, 
	%and for video tasks  include representation learning~\cite{videobert,mmt,CBT}, and action classification~\cite{nonlocal,actiontfm,featbank}.
	
	\paragraph{Visual Question and Answering (VQA).} Models for VQA usually have queries which attend to relevant features for predicting the answers~\cite{focal,multiturn,multimodal,learnable}. For example,~\cite{tgif} use co-attention between vision and language, and~\cite{attr_attn} adapts attribute-based attention in LSTM using a pertained attribute detector.~\cite{progressive} proposes a progressive attention memory to progressively prune out irrelevant temporal parts. Our query decoder has a similar query-response mechanism, however, our final goal is action recognition not VQA. Instead of having specific questions for each video, we are interested in a common set of queries shared across the whole dataset.
	
	%** need to review VQA, especially for videos, and then contrast
	%with what we are doing. VQA typically casts the problem as a text query, and the response
	%has a multiple choice for each query. Also, usually the query is for a particular image or video clip, i.e.\ not untrimmed videos ** 
	
	%% TO CITE:
	% \cite{sener2020temporal} -- not directly related, but cite.
	
		\begin{figure}[tbp]
		\centering
		\includegraphics[width=0.48\textwidth]{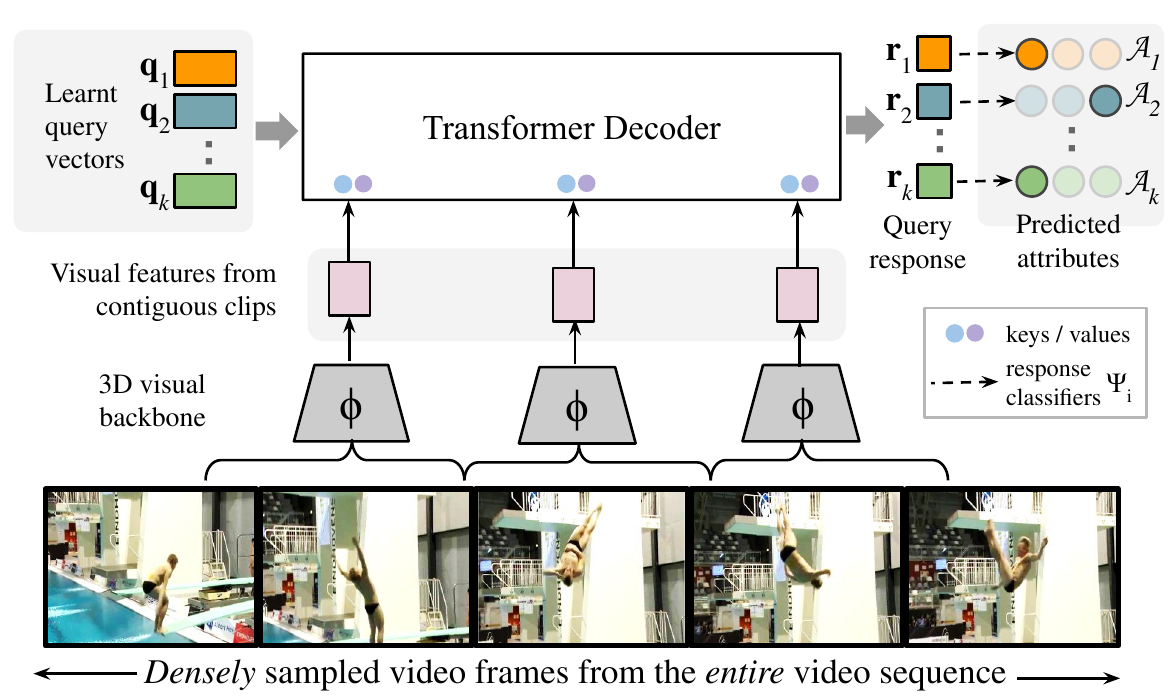}
		\caption{\textbf{Temporal Query Network.} 
			% \TQN{}s answer detailed questions about event types and attributes for videos. 
			A set of permutation-invariant {\em query vectors} $\bq_i$ are learnt
			for pre-defined queries. They attend over densely extracted visual features in a Transformer~\cite{vaswani2017attention} decoder
			 and generate  {\em response vectors} $\br_i$, which are
			 linearly classified (${\bf\Psi}_i$) into 
			attributes $a_i^j$ from corresponding attribute sets $\mathcal{A}_i$. }\label{f:arch}
	\end{figure}
	
	\section{Method}\label{sec:method}
	In this section we first, describe the {\em Temporal Query Network} (\TQN) 
	decoder, which given only weak supervision (no event location/duration), learns 
	to respond to the queries by attending over the entire {\em densely} sampled 
	untrimmed video (\Cref{sec:tqn-model}). Second, we introduce a training 
	framework to overcome GPU memory constraints preventing use of temporally-dense 
	video input (\Cref{sec:ft-bank}). Finally, we explain how the monolithic 
	category labels that are normally provided with fine-grained video datasets, and 
	typically composed of a varying number of sub-labels (or tokens drawn from a 
	finite-set) corresponding to event types and attributes, can be factored into a 
	set of queries and corresponding attributes (\Cref{sec:factor}).
	
	\subsection{Temporal Query Networks}\label{sec:tqn-model}
	A {\em Temporal Query Network} (\TQN) identifies rapidly occurring discriminative 
	events (spanning only a few frames) in untrimmed videos, and can be trained given 
	only weak supervision, \ie, no temporal location or duration information for 
	events. It achieves this by learning a set of permutation-invariant {\em query 
		vectors} corresponding to pre-defined queries about events and their attributes,
	which are transformed into {\em response vectors} using 
	Transformer~\cite{vaswani2017attention} decoder layers attending to visual 
	features extracted from a 3D ConvNet backbone. \Cref{f:arch} gives an overview 
	of the model. The visual backbone and the \TQN decoder are described below. 
	
	\paragraph{Query--Attributes.} The query set is $\mathcal{Q}~=~\{q_i\}_{i=1}^K$, 
	where each query $q_i$ has a corresponding attribute set 
	$\mathcal{A}_i~=~\{a^i_1,a^i_2,\hdots,a^i_{n_i-1},\varnothing\}$ 
	consisting of the admissible values $a^i_j$ in response to $q_i$; $\varnothing$ 
	denotes the {\tt null} value (not present), and the total number 
	of attributes $n_i = |\mathcal{A}_i|$ is query dependent.
	
	For example, in diving videos a query could be the {\tt number of turns} with
	the attribute set being the possible counts ${\tt\{0.5, 1.0, 2.5\}}$; or in 
	gymnastics, the query could be the {\tt event type} with attributes {\{\tt vault, floor-exercise,
		balanced beam\}}.

	\paragraph{Visual backbone.} Given an untrimmed video, first visual features for
	contiguous non-overlapping clips of 8 frames are extracted using a 
	3D ConvNet: ${\bf \Phi} = (\Phi_1,\Phi_2,\hdots,\Phi_t)$, where $t$ is the total 
	number of clips, and $\Phi_i\in\mathbb{R}^d$ is the $d$-dimensional clip-level 
	visual feature.  Note, it is important to extract features {\em densely} from 
	the {\em entire} length of the video because: (i)~it avoids causing temporal 
	aliasing and also missing rapid events (which span only a few frames), \eg, a 
	somersault, and (ii)~selecting a subset of clips from the full video for 
	classification~\cite{i3d,s3d,x3d} is sub-optimal as the location 
	of these events is unknown.
	
	\paragraph{TQN Decoder.} Given the clip-level features and the label queries,
	the \TQN decoder outputs a {\em response} for each query. Concretely, for each 
	label query $q_i$, a vector $\bq_i \in \mathbb{R}^{d_q}$ is learnt for which a 
	response vector $\br_i \in \mathbb{R}^{d_q}$ is generated by attending over the 
	visual features ${\bf \Phi}$. Each response vector $\br_i$ is then linearly 
	classified independently into the corresponding attribute set $\mathcal{A}_i$. 
	
	In more detail, we use multiple layers of a {\em parallel non-autoregressive} 
	Transformer decoder, as also used in~\cite{detr}. 
	Each decoder layer first performs self-attention between the queries, followed 
	by multi-head attention between the updated queries and the visual features.
	% to output the response vectors.
	In each attention head, the visual features ${\bf\Phi}$ are used to linearly 
	regress {\em keys} $\Gamma\cdot{\bf\Phi}$ and {\em values} 
	$\Lambda\cdot{\bf\Phi}$, where $\Gamma$ and $\Lambda$ are the linear key and
	value heads. The values are gathered using Softmaxed dot-products between the 
	keys and queries as the weights. Finally, a feed-forward network ingests the 
	values from multiple heads and outputs the response vectors. 
	The response vectors from one decoder layer act as queries for the next layer, except for the
	first layer where the learnt queries $\bq$ are input. Hence, each decoder layer
	refines the previously generated response vectors. Mathematically, if 
	$\ell^{(j)}$ is the $j${th} decoder layer, $j\in \{1,2,\hdots,M\}$:
	\begin{equation}
	\begin{split}
	\setlength{\abovedisplayskip}{4pt}
	\setlength{\belowdisplayskip}{4pt}
	& \ell^{(j)}(\cdot,\cdot): \mathbb{R}^{N\times d_q}\times\mathbb{R}^{t\times d} \mapsto \mathbb{R}^{N\times d_q} ,\\
	& \ell^{(j)}  ({\bf r}^{(j-1)},  {\bf \Phi}) \mapsto {\bf r}^{(j)},\quad \text{and}  \\
	& {\bf r}^{(0)} \triangleq {\bf q}.
	\end{split}
	\end{equation}
	The response vectors from the final ($M$th) layer $\br^{(M)}_i \in 
	\mathbb{R}^{d_q}$ corresponding to the queries $\bq_i$, $i \in \{1, 2, \hdots, 
	K\}$ are classified into the corresponding attribute sets $\mathcal{A}_i$ 
	using $K$ independent linear classifiers $\Psi_i: \mathbb{R}^{d_q} \mapsto 
	\mathbb{R}^{n_i}$, where $n_i$ is the query dependent total number of admissible 
	attributes. Please refer to \Cref{f:arch} for a visual representation of this 
	process, and \sref{s:supp-impl} for details of the
	Transformer decoder. 
	
	\paragraph{Training.} The model parameters, \ie, from the visual encoder 
	and the \TQN decoder are trained jointly end-to-end with the attribute 
	classifiers $\Psi_i$ through backpropagation. The training loss is a multi-task 
	combination of individual classifier losses, which are
	Softmax cross-entropy $\mathcal{L}_{CE}$ losses on the logits 
	$\Psi_i\cdot\br^{(M)}_i$ over the attribute sets $\mathcal{A}_i$:
	\begin{equation}
	\setlength{\abovedisplayskip}{4pt}
	\setlength{\belowdisplayskip}{4pt}
	\mathcal{L}_{\text{total}} = \sum_{i=1}^K \mathcal{L}_{CE}^{(i)}(a^i, 
	\Psi_i\cdot{\bf r}^{(M)}_i),
	\end{equation}
	where $a^i$ is the ground-truth attribute for the label query $q_i$.
	
\begin{figure*}
		\includegraphics[width=\linewidth]{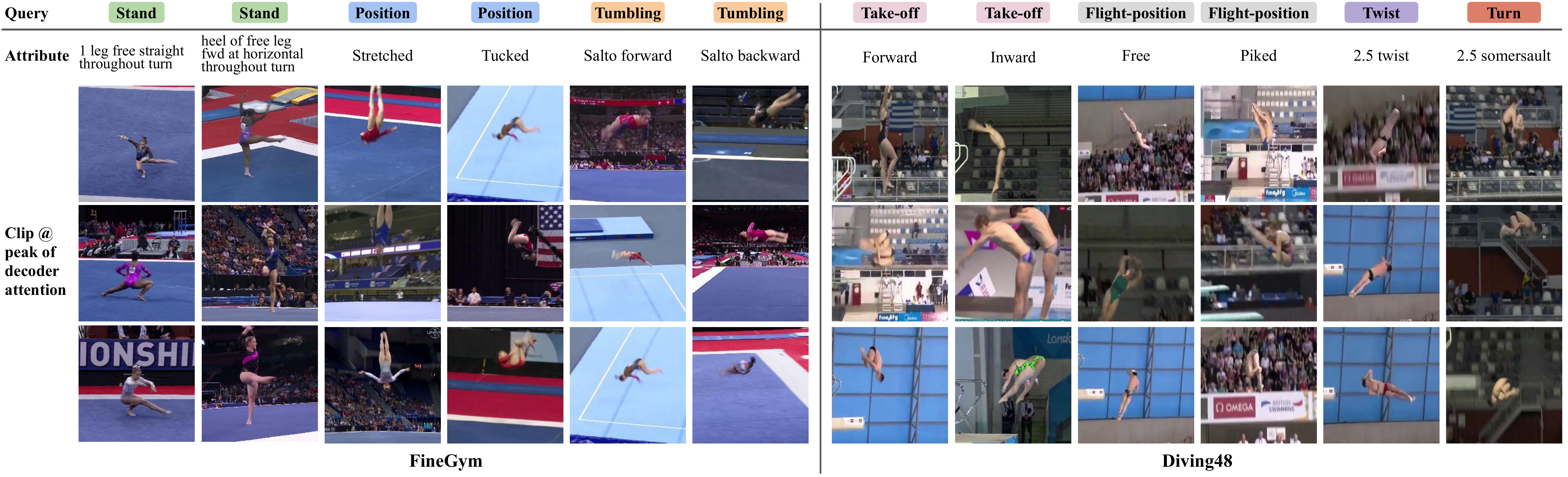}%
		\vspace{-1mm}%
		\caption{{\bf TQN attention alignment.}  
			For a given query, the \TQN attends over the clip-level features to generate
			the responses. We visualize the central frame from the clip with the highest
			attention score for six different query-attributes from the 
			FineGym~\cite{shao2020finegym} dataset.
			The same queries (but different attributes) are highlighted with a 
			common color. The \TQN detects and aligns semantically relevant 
			events under variations in appearance and pose without any
			temporal localization supervision. More visualizations in .}%
		\label{f:attnviz-peak-clip}
	\end{figure*}

\begin{figure*}
		\includegraphics[width=\linewidth]{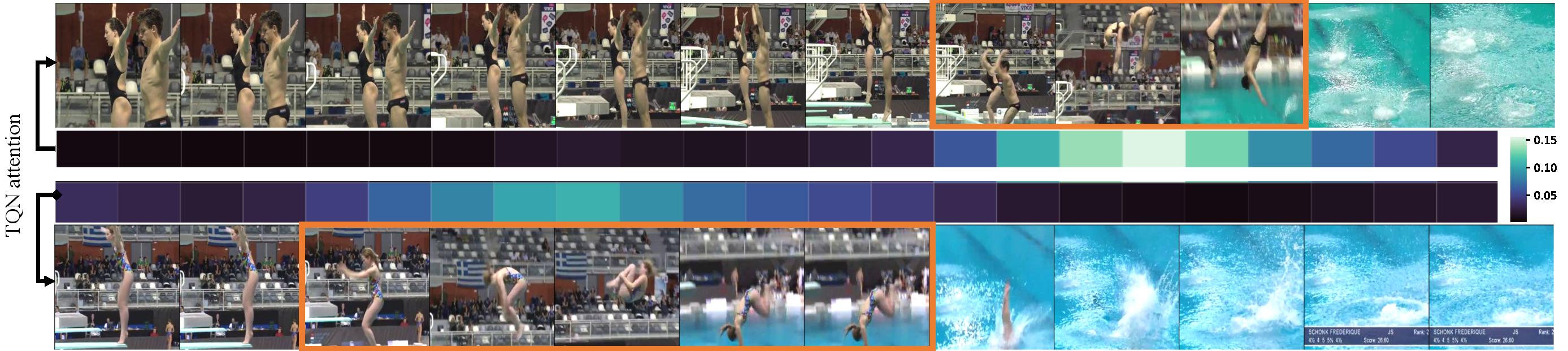}
		\caption{{\bf TQN temporal attention.} \textbf{{\color{blue} Blue}} colored maps visualize the attention averaged over all queries predicted by \TQN for two clips from the Diving48 dataset~\cite{li2018resound}. The peaks in these maps correspond to temporal location of diving `flight' highlighted in \textbf{{\color{orange} orange}}. \TQN rejects non-informative frames at the start and end of untrimmed videos to localize discriminative 
			frames relevant for fine-grained recognition.}\label{f:attnviz-time}
		\vspace{-3mm}
	\end{figure*}

	In essence, the \TQN decoder learns to establish \emph{temporal correspondence}
	between the query vectors and the relevant visual features to generate the 
	response. Since the query vectors are themselves learnt, they are optimized 
	to become `experts' which can localize the corresponding event in the 
	untrimmed temporal feature stream.
	\Cref{f:attnviz-peak-clip,f:attnviz-time} illustrate this temporal
	correspondence. 

	\paragraph{Discussion: TQN and DETR.}
	DETR~\cite{detr} is a recently proposed Transformer based object detection 
	model, which also similarly employs non-autoregressive parallel decoding to 
	output object detections at once. However, there are three crucial differences: 
	(i)~the DETR object queries are all equivalent -- in that their outputs all specify 
	the same `label space' (object classes and their RoI), essentially queries are 
	{\em learnt} position encodings. In contrast the \TQN queries are distinct from 
	each other and carry a semantic meaning corresponding to event types and 
	attributes; their output response vectors each specify a different set of 
	attributes, and the number of attributes is query dependent. 
	(ii) This leads to the second difference: 
	since the \TQN responses are tied to these queries, they can be trained with 
	direct supervision for attribute labels, thereby avoiding train-time Hungarian 
	Matching~\cite{kuhn1955hungarian} between prediction and ground-truth employed 
	in DETR. 
	(iii)~Finally, no temporal localization supervision is available to the \TQN,
	while (spatial) locations are provided for DETR training. Hence, although \TQN
	is tasked with (implicit) detection of events, it does so with much weaker 
	supervision.

	\begin{figure}[t]
		\includegraphics[width=\linewidth]{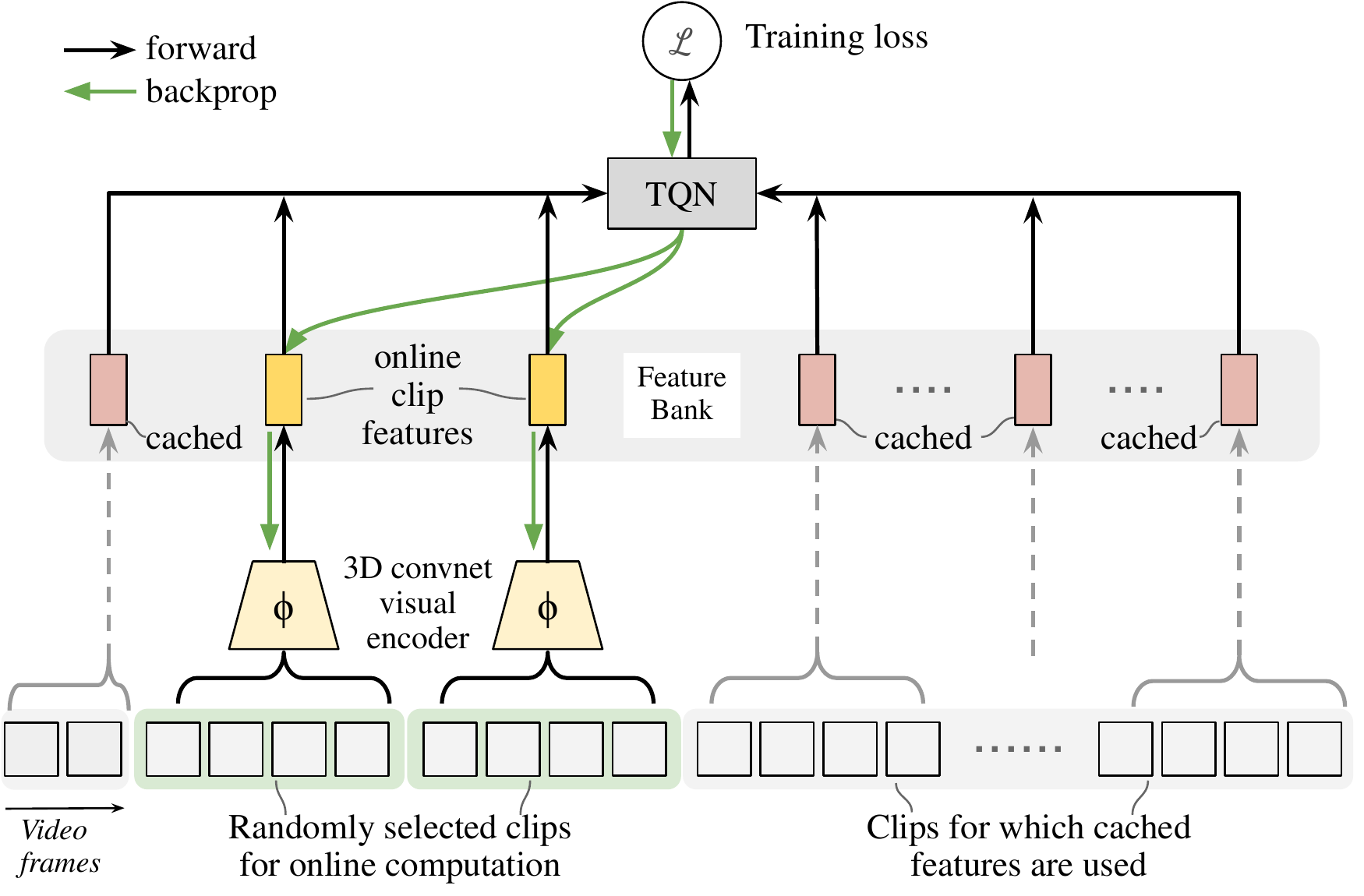}
		\caption{{\bf Stochastically updated feature bank.} Feature banks cache 
			visual encoder features $\Phi$ to circumvent GPU memory constraints which
			prevent forwarding densely sampled video frames from the entire length of 
			the video at each training iteration. Randomly sampled contiguous video clips 
			are forwarded online in each iteration and cached immediately; the rest of 
			the clip features are retrieved from the feature bank. The features are 
			then input into the \TQN for {\em joint} training of both the  \TQN and the visual 
			encoder. This joint training over dense temporal context is critical for 
			fine-grained performance (\Cref{s:ft-bank-ablation}).}\label{f:ft-bank}
	\end{figure}
	\subsection{Stochastically Updated Feature Bank}\label{sec:ft-bank}
	Dense temporal sampling of frames for the entire untrimmed video input is key 
	for detecting rapid discriminative events with unknown
	temporal location. However, this is challenging in practice due to GPU memory
	constraints which prevent forwarding densely sampled frames in each training
	iteration. We use a feature memory bank~\cite{xiao2017joint,wu2018unsupervised} 
	to overcome these constraints.
	
	The memory bank caches the clip-level 3D ConvNet visual features.
	Note for a given video, the clip features ${\bf\Phi} = (\Phi_i, \Phi_2, 
	\hdots,\Phi_t)$, where $t$ is the total number of clips, can be extracted 
	independently of each other. 
	The memory bank is initialized with clip features for all the training 
	videos extracted from a pre-trained 3D ConvNet (details in \Cref{s:impl}).
	Then in each training iteration, a fixed number $n_{\text{online}}$ of 
	randomly sampled consecutive clips are forwarded through the visual encoder, 
	\ie, $n_{\text{online}}$ clip features are computed {\em online}.
	The remaining $(t - n_{\text{online}})$ clip features are retrieved from the 
	memory bank. % and act as constants
	The two sets of visual features are then 
	combined and input into the \TQN decoder for final prediction and 
	backpropagation to update the model  parameters. 
	Finally, the clip features in the memory bank corresponding to the ones computed 
	online are replaced with the online features. During inference, all the features 
	are computed online without the memory bank.
	\Cref{f:ft-bank} summarizes the function visually.
	
	\paragraph{Advantages.} Using a fixed number of clips online decouples the 
	length of videos and the GPU memory budget. As a result our memory 
	bank enables the \TQN decoder to be trained (i)~jointly with the visual encoder, 
	(ii)~with extended temporal context, both of which impart drastic improvements 
	in performance (see \Cref{s:exp-mbank}).
	Further, it promotes diversity in each mini-batch as multiple different videos 
	can be included instead of just a single long video.
	
	\paragraph{Discussion: relations with prior memory bank methods.}
	Feature memory banks have been used for compact vector representations of single 
	images~\cite{xiao2017joint, wu2018unsupervised, tian2019contrastive}, 
	whereas we store a varying number of temporal vectors for each video.
	MoCo~\cite{moco} and related self-supervised methods~\cite{wang2020self, 
		yang2020vthcl} update the memory bank features {\em slowly} \eg, using a 
	secondary network, to prevent representation collapse, whereas given 
	direct supervision for the queries, we can update the features immediately 
	from the single online network.
	The above works apply memory banks for image/feature retrieval from a large 
	corpus for hard negatives in contrastive training, while we use the memory bank for extending 
	the temporal context for each video. 
	Using pre-computed features~\cite{li2017temporal, miech2017learnable, 
		tang2018non} or {\em Long-term Feature Banks}~\cite{featbank} are 
	prominent~\cite{beery2020context, garcia2020knowledge, wu2020context} strategies
	for extending the temporal support of video models.
	However, all of these works keep the features {\em frozen}, while we 
	{\em continuously~update} the memory-bank during training. In 
	\Cref{s:exp-mbank}, we demonstrate these updates are critical for performance.

	\begin{table}[b]
		\setlength{\tabcolsep}{4pt}
		\resizebox{\linewidth}{!}{%
			\begin{tabular}{ll|cc|ccc}
				\hline
				\multicolumn{1}{c|}{\multirow{2}{*}{\textbf{Category} $\downarrow$}} &\multicolumn{1}{r|}{\textbf{query} $\rightarrow$ }   & \multicolumn{2}{c|}{$q_1$: leap and jump type} & \multicolumn{3}{c}{$q_2$: num turns} \\ \cline{3-7} 
				& \multicolumn{1}{|r|}{\textbf{attribute}$\rightarrow$} & switch leap            & split jump            & 0.5      & 1.0      & $\varnothing$   \\ \hline
				\multicolumn{2}{l|}{switch leap w/ 0.5 turn}                              & \cmark                 &                       & \cmark   &          &                 \\
				\multicolumn{2}{l|}{switch leap w/ 1 turn}                                & \cmark                 &                       &          & \cmark   &                 \\
				\multicolumn{2}{l|}{split jump w/ 1 turn}                                 &                        & \cmark                &          & \cmark   &                 \\
				\multicolumn{2}{l|}{split jump}                                           &                        & \cmark                &          &          & \cmark          \\ \hline
		\end{tabular}}
		\vspace{1mm}
		\caption{{\bf Illustration of query-attribute factorization of fine-grained 
				action categories.} Four categories are factored into two queries
			$q_1,q_2$ with two and three attributes respectively.}\label{t:factor}
	\end{table}

	\subsection{Factorizing Categories into Attribute Queries}\label{sec:factor}
	
	In this section we illustrate how the 
	% The objective in fine-grained video recognition is to classify a given video 
	% $\bf x\in\mathbb{R}^{T\times H\times W\times 3}$ of $T$ RGB 
	% $H{\times} W$ sized frames, 
	pre-defined set of $N$ categories 
	$\mathcal{C}~=~\{c_1, c_2, \hdots, c_N\}$ typically associated with fine-grained video recognition datasets can
	be factored into attribute queries.
	In such datasets, the categories differ in subtle details \eg, the specific type,
	duration, or count of a certain sequence of events. These can be rapidly 
	occurring (short duration) events with unknown temporal location and duration 
	(see \Cref{f:teaser}).  
	
	% Each category label is composed of multi-part textual description of events and their attributes. 
	The textual descriptions of categories $\mathcal{C}$ are strings composed of a 
	varying number of sub-labels (or tokens drawn from a finite-set) corresponding 
	to event types and attributes (sub-label categories). We leverage this string 
	structure to form queries $\{q_i\}_{i=1}^K$ corresponding to sub-label categories,
	each with an associated attribute set $\mathcal{A}_i$ composed of sub-labels,
	such that the categories $\mathcal{C}$ can be expressed as a subset of the 
	cartesian product of the attribute sets: 
	$\mathcal{C}\subseteq\mathcal{A}_1\times\mathcal{A}_2\times\hdots\times\mathcal{A}_K$. 
	An example label factorization for four categories is given in 
	\Cref{t:factor}, where four action categories are expressed as a product of two 
	attribute sets containing two and three attributes respectively. 
	Factorization details for the evaluation datasets used in this paper are given 
	in \Cref{s:data} and \sref{s:label-factor-list}.
	
	%%%% The following is not what we do, as we use a "global" class query for prediction
	%%%% So have taken it out.
	% Given a \TQN-style {\em query}-based video parsing model trained on above
	% query attributes, the final prediction into categories $\mathcal{C}$ is obtained
	% as the $j$th $c_j$ category with the highest joint probability of its constituent 
	% attributes $(a^{j,1},a^{j,2},\hdots,a^{j,K})$,
	% % \begin{equation}
	%     $c_{\text{pred}} = \underset{j}{\operatorname{argmax}} \enskip 
	%     \prod_{i=1}^K\enskip p(a^{j,i})$.
	% % \end{equation}
	
	This factorization unpacks the monolithic category labels into their 
	semantic constituents (the queries and attributes).
	It improves data-efficiency, through sharing video data across common sub-labels 
	(instead of disjoint category-specific data), and induces \TQN-style
	query-based temporal localization and classification video parsing models.

	\subsection{Implementation Details}\label{s:impl}
	We describe key model and training details below, with further details in 	\sref{s:supp-impl}.
	
	\paragraph{Model architecture.}%
	We use S3D~\cite{s3d} as visual backbone, operating on non-overlapping
	contiguous video clips of 8 
	frames each of size $224\times 224$ pixels  with consistent temporal stride $s$ ($s$=1 in FineGym, $s$=2 in Diving48), to output one feature vector per clip.
	The decoder consists of four standard 
	post-normalization~\cite{xiong2020layer} Transformer decoder
	layers~\cite{vaswani2017attention}, each with four attention heads, and 1024-dim keys, (learnt-) 
	queries, and values. Dropout rate is 0.1 in the decoder and 0.5 for output features.
	
	\paragraph{Training.}\label{s:two-stage}
	The visual encoder is pre-trained on Kinetics-400~\cite{kinetics}. 
	Then, we proceed by a two stage curriculum. First, the model is trained {\em end-to-end} on 
	{\em short} videos containing fewer than $K$ frames (FineGym: $K=48$, 
	Diving48: $K=128$; as the latter contains approx.\  3$\times$ longer videos), such 
	that they can fit on two Nvidia RTX 6000 GPUs with batch size 16. Second, the model is 
	trained on the whole training set using the stochastically updated memory bank 
	(\Cref{sec:ft-bank}) to accommodate long videos. We use the Adam 
	optimizer~\cite{Kingma15}, and train for 50 epochs in the first stage, followed 
	by 30 more epochs in the second.
	
	\section{Datasets, Baselines, Label Sets, and Metrics}\label{s:data}
	
	We evaluate \TQN for fine-grained action recognition on two video datasets, 
	namely, FineGym~\cite{shao2020finegym}, and Diving48~\cite{li2018resound}.
	We introduce the datasets, list the baselines methods we compare against,
	detail the query-attribute based label sets for them, and
	state the evaluation measure below.
	
	\paragraph{FineGym.}%
	FineGym is a recently introduced dataset (2020) for fine-grained action understanding, 
	consisting of HD gymnasium classes with subtle motion details.
	% In more detail, categories are labelled at
	% three semantic levels, namely (from coarse to fine)---{\em event}, {\em set}, 
	% and {\em element}, and temporal location for actions and sub-actions is provided.
	We evaluate for classification under two settings specified in the dataset with standard train/eval splits: 
	(1)~Gym99: relatively balanced data for 99 categories with
	26k training/8.5k testing videos; and (2)~Gym288: long-tailed data for 288
	categories with 29k training/9.6k testing videos. 
	There is a large variation in video lengths: min: 13 frames, max: 877 frames, average: 47 frames.
	
	\paragraph{Diving48.}%
	Diving48 contains competitive diving video clips from 48 classes.
	It similarly evaluates fine-grained video recognition by having a common diving 
	setting where subtle details of diving sequences define the various categories 
	instead of coarse objects or scenes.
	We use the standard split containing 16k training/2k test videos.
	The video lengths have a very wide range: min: 24 frames, max: 822 frames, average: 158 frames. We use the \emph{cleaned-up} labels (denoted `v2') released in
	Oct 2020. Results for the noisy version (Diving48-v1) can be found in \sref{s:noisy-diving}.
	
%	\paragraph{Noisy labels and Diving48-v2.}
%	The Diving48 authors recently posted cleaned-up labels for the 
%	dataset\footnote{The updated Diving48 labels posted on 30 Oct 2020 are denoted `V2' on 
%		the webpage: 
%		{\urlstyle{same} \url{http://www.svcl.ucsd.edu/projects/resound/dataset.html}}.}.
%	The difference between the previous (v1) and new labels (v2) reveals that {\em 43.5\%
%		of training, and 35.8\% of test videos were mislabelled in v1}, rendering
%	all previously reported results difficult to interpret.
%	In order to compare to previous work on v2,  we trained publicly available SotA methods (listed below) on v2 and
%	report the results in \Cref{s:exp-sota}. For 
%	completeness, we also compare against `SotA' on the original noisy (v1) labels in 
%	\sref{s:noisy-diving}, even though this comparison is not informative due to
%	heavy label noise.
	
	\paragraph{Diving48-v2 SotA comparison.} We trained publicly available SotA action 
	recognition models on v2 labels, namely: (i)~TSM~\cite{tsm}, (ii)~TSN~\cite{tsn},
	(iii)~TRNms~\cite{trn}, (iv)~I3D~\cite{i3d}, (v)~S3D~\cite{s3d}, and 
	(vi)~GST-50~\cite{gst}. 
	Note, GST-50 is the top-performing method on Diving48-v1 amongst those using a 
	ResNet-50 backbone (refer to v1 comparison in \sref{s:noisy-diving}).
	The original dataset paper~\cite{li2018resound} reports results only for TSN and 
	C3D~\cite{c3d}; C3D (2013) is omitted as it is outperformed by more recent 
	methods, \eg, I3D and S3D. We could not benchmark against other prominent 
	methods reported for v1, namely, CorrNet~\cite{corrnet} and 
	AttnLSTM~\cite{kanojia2019attentive}, as their implementation is not public.
	
	\paragraph{Baseline methods.}
	Since we use S3D~\cite{s3d} as the visual backbone for \TQN, the following two 
	methods form the baselines:
	(i)~\textbf{Short-term S3D (ST-S3D):} following 
	the original dataset papers~\cite{shao2020finegym,li2018resound}, it is 
	trained on single clips of fixed number (=8 if not specified otherwise) of 
	frames, while for inference,
	the predicted probabilities from multiple clips spanning a given video are
	averaged for final classification.
	(ii)~\textbf{Long-term S3D (LT-S3D = S3D + Feature Bank):} uses our \SUFB at 
	training time in order
	to pool information from the entire video. Specifically,
	LT-S3D replicates the ST-S3D's multi-clip evaluation setting at training time, 
	\ie, class probabilities obtained from multiple clips spanning the entire 
	video are averaged and used for prediction and backpropagation.
	
	Note at training time ST-S3D incorrectly bases its decisions on {\em individual} 
	clips which may not contain information relevant for classification. 
	LT-S3D overcomes this issue using multi-clip \sufb and learns better clip-level
	features (\Cref{s:ft-bank-ablation}).
	
	% trained on {\em individual} clips, hence its decisions are
	% on correlation amongst temporally-local short clips, which may not contain
	% the content relevant for classifying the video, while
	% LT-S3D's prediction at training time is based on multiple-clips, and hence, 
	% it learn to spot clips relevant for the final classification.
	% It is evaluated in the same way as in ST-S3D, but in training we use our 
	% \SUFB so that the network makes decision using information from $N$ clips --  
	% both online computed features and offline features stored in  the \SUFB  -- 
	% to close the gap between training and inference time.  
	% During inference, $N$ clips are densely sampled from a given video (
	% 	$N$ proportional to the length of the video), the class probabilities from 
	% 	these clips are averaged for final classification. 
	% This approach is used 
	% widely in video classification but not suitable for this task as it is 
	% trying to classify the whole video by looking at short clips without the 
	% context.

%	 ** Is the input to S3D always 8-frames? It's not clear from this description. Add another sentence here explaining what this means, or give an example. **

	\paragraph{Label sets: query-attributes.}
	We define 
	query-attributes independently for each dataset. 
	\textbf{Diving48:} The original 48 classes are defined in terms of four 
	{\em stages} of a diving action.
	We use four queries corresponding to the stages, and the possible instantiation of each stage as attributes. 
	\textbf{FineGym:} Each category in Gym99 and Gym288 is defined by a textual 
	description for the specific sequence {\em elements} in a gymnastic set, \eg,
	{\em ``double salto backward tucked with 2 twist''}.
	We extract nouns from these and categorize them into 12 queries, \eg, {\em swing, landing, 
	jump and leap,} \etc.  and their instantiations form the attributes.
	Complete factorization is specified in \sref{s:label-factor-list}.
	In addition to these query and attribute sets, we augment
	the \TQN query set with a ``global''query class with the original fine-grained 
	categories as its attribute set, and use its response for the final category 
	prediction.%\ag{check the sensitivity of performance}

	\paragraph{Metrics.} We evaluate on (top-1) classification accuracy, both 
	per original class (48 in Diving48, 99 in Gym99 and 288 in Gym288), and per video.
	
	\begin{table}[t]
		% \vspace{-3mm}
		\renewcommand{\arraystretch}{1.4}
		\resizebox{\linewidth}{!}{
			\begin{tabular}{c|c|c|c|c|c|c}
				\hline
				\multirow{2}{*}{\bf Backbone} & \multirow{2}{*}{\bf Encoder}        & \multirow{2}{*}{\begin{tabular}[c]{@{}c@{}}{\bf Decoder}\\ (Aggregation)\end{tabular}} & \multirow{2}{*}{\bf Classification}                                                        & \multirow{2}{*}{\bf Label}                                                        & \multicolumn{2}{c}{\bf Accuracy} \\ \cline{6-7} 
				&                                 &                                                                                  &                                                                                        &                                                                               & {\bf per-class}     & {\bf per-video}     \\ \hline
				\multirow{5}{*}{S3D}      & --                               & average pooling                                                                  & \multirow{2}{*}{\begin{tabular}[c]{@{}c@{}}multi-class\\ (cross entropy)\end{tabular}} & \multirow{2}{*}{class index}                                                  & 72.3          & 80.4          \\ \cline{2-3} \cline{6-7} 
				& \multirow{3}{*}{self-attention} & --                                                                                &                                                                                        &                                                                               & 73.7          & 80.0          \\ \cline{3-7} 
				&                                 & --                                                                                & \begin{tabular}[c]{@{}c@{}}multi-label\\ (binary cross entropy)\end{tabular}           & \multirow{3}{*}{\begin{tabular}[c]{@{}c@{}}text \\ descriptions\end{tabular}} & 47.9          & 50.3          \\ \cline{3-4} \cline{6-7} 
				&                                 & \begin{tabular}[c]{@{}c@{}}auto-regressive  \\ Transformer\end{tabular}  & \begin{tabular}[c]{@{}c@{}}sequence prediction\\ (cross entropy)\end{tabular}          &                                                                               & 51.9          & 65.1          \\ \cline{2-4} \cline{6-7} 
				& --                               & TQN                                                                              & \begin{tabular}[c]{@{}c@{}}multi-task\\ (cross entropy)\end{tabular}                   &                                                                               & \textbf{74.5 }         & \textbf{81.8  }        \\ \hline
			\end{tabular}%
		}
		\vspace{1mm}
		\caption{\textbf{Leveraging multi-part text descriptions.}  
			We compare our {\em query-attribute} label factorization (\Cref{sec:factor})
			to alternative methods for learning with unaligned (no temporal location 
			information) multi-part text descriptions.
			Our \TQN + label factorization 
			outperforms other approaches which are representative of standard classification, 
			and modern encoder-decoder architectures for sequences (see \Cref{s:multi-att-lbl}).
			Evaluation on the Diving48-v2 dataset.}
		\label{tab:texts}
	\end{table}

	\begin{table*}[t]
		\centering
		\resizebox{\textwidth}{!}{%
			\begin{tabular}{cccc|cccc|cc|cc}
				\hline
				\multirow{3}{*}{\textbf{Network}} & \multirow{3}{*}{\textbf{Pretrained dataset}} & \multirow{3}{*}{\textbf{Modality}} & \multirow{3}{*}{\textbf{\# frames in training}} & \multicolumn{4}{c|}{\textbf{Gym99}}                         & \multicolumn{2}{c|}{\textbf{Gym288}}                     & \multicolumn{2}{c}{\textbf{Diving48-v2}}                  \\ \cline{5-12} 
				&                                              &                                    &                                                 & \multirow{2}{*}{Per-class} & \multicolumn{3}{c|}{Per-video} & \multirow{2}{*}{per-class} & \multirow{2}{*}{per-video} & \multirow{2}{*}{per-class} & \multirow{2}{*}{per-video} \\ \cline{6-8}
				&                                              &                                    &                                                 &                            & subset VT  & subset FX & total &                            &                            &                            &                            \\ \hline
				I3D                               & K400                                         & RGB                                & 8                                               & 64.4                       & 47.8       & 60.2      & 75.6  & 28.2                       & 66.7                       & 33.2                       & 48.3                       \\
				TSN                               & ImageNet                                     & two-stream                         & 3                                               & 79.8                       & 47.5       & 84.6      & 86    & 37.6                       & 79.9                       & 34.8                       & 52.5                       \\
				TSM                               & ImageNet                                     & two-stream                         & 3                                               & 81.2                       & 44.8       & 84.9      & 88.4  & 46.5                       & 83.1                       & 32.7                       & 51.1                       \\
				TRNms                             & ImageNet                                     & two-stream                         & 3                                               & 80.2                       & 47.3       & 84.9      & 87.8  & 43.3                       & 82.0                       & 54.4                       & 66.0                    \\
				GST-50                               & ImageNet                                     & RGB                                & 8                                               & 84.6                       & 53.6       & 84.9      & 89.5  & 46.9                       & 83.8                       & 69.5                       & 78.9                       \\
				ST-S3D                            & K400                                         & RGB                                & 8                                               & 72.9                       & 45.3       & 82.8      & 81.5  & 42.4                       & 75.8                       & 36.3                       & 50.6                       \\ \hline
				LT-S3D                            & K400                                         & RGB                                & dense                                           & 88.9                       & 69.1       & 90.4      & 92.5  & 57.9                       & 86.3                       & 72.3                       & 80.4                       \\
				TQN                               & K400                                         & RGB                                & dense                                           & \textbf{90.6   }                    & \textbf{74.9}       & \textbf{91.6 }     & \textbf{93.8}  & \textbf{61.9  }                     & \textbf{89.6   }                    & \textbf{74.5  }                     & \textbf{81.8 }                      \\ \hline
			\end{tabular}%
			
		}\vspace{1mm}%
		\caption{\textbf{Comparison to state-of-the-art.} We compare \TQN to several 
			SotA methods for Gym99, Gym288, and Diving48-v2. The results for the Gym 
			datasets are reproduced from the original dataset 
			publication~\cite{shao2020finegym}, except for S3D~\cite{s3d} and GST-50~\cite{gst} was trained by us; 
			no further results are available as the 
			dataset was recently published (2020). For Diving48-v2, since the corrected 
			labels were released recently, we train the publicly available 
			implementations of all methods while replicating the setting of their 
			application to the original Diving48-v1 dataset. \TQN achieves 
			top-performance on all three datasets, detailed discussion in \Cref{s:exp-sota}.}
		\label{t:sota}\vspace{-4mm}
	\end{table*}

	% We report top-1 accuracy per class and top-1 accuracy per video.
	
	% \paragraph{Class Accuracy}We explain the methods we use to map different model output to class predictions for accuracy computation.
	% For baseline models where no textual information is used, we train a linear classifier on the output from the model to predict the probabilities of classes.
	% For models which output a fixed number of tokens from each video (e.g, TQN), we map the predicted tokens probabilities  $P_{\text{token}} \in \mathbb{R}^{ L \times  K}$ from one video to the class probabilities $P_{\text{class}} \in \mathbb{R}^{N}$, where $L$ is the length of token sequence, $K$ is the size of tokens in the dataset,  $N$ is the number of classes.
	% The probability of certain class $P_{\text{class}}^{(i)}$, is aggregated by: 
	% \begin{equation}
	% P_{\text{class}}^{(i)} = \prod \limits_{j=1}^L {P_{{\text{token}}}^{(j)}*y_{i}^{(j)} } 
	% \label{e:class_prob}  
	% \end{equation}
	% where $y_{i}^{(j)}  \in \mathbb{R}^{1 \times  K}$ is the ground truth one-hot vector at position $j$ in class $i$.

	\section{Experiments}\label{s:exps}
	% In \cref{s:exp-ablation} we conduct a thorough ablation study on memory feature bank and architecture of query decoder. We first explore and compare different methods of doing fine-grained action recognition, with or without supervision from texts. We then look into the importance of temporal support in training and test time, and investigate the possible of providing full temporal support in training, followed by evaluation on the gap between using stochastic memory feature bank and end-to-end training.
	% Finally, In \cref{s:exp-sota} we compare our best TQN model with the State-of-the-Art results from other models.
	
	% ** add a short overview of the the contents if room **
	
	\subsection{Leveraging Multi-Attribute Labels}\label{s:multi-att-lbl}
	We evaluate alternative methods and losses for exploiting multi-part text 
	descriptions on the Diving48-v2 dataset, and compare performance to our
	{\em query-attribute} label factorization (\Cref{sec:factor}).
	\Cref{tab:texts} summarizes the results for various encoders, decoders and losses;
	detailed description are given in \sref{s:text-sup}.
	\TQN multi-task losses perform the best (81.8\% per-video
	accuracy), followed by standard multi-class classification (avg.\ pool: 
	80.4\%, self-attention: 80.0\%) which only uses the class index, not the text 
	descriptions.
	Other sequence based methods for text perform substantially worse ($-$15-30\%),
	due to the restrictive ordering imposed by the text string.
	\TQN goes beyond just attention-based context aggregation, as it 
	outperforms S3D+attention-encoder trained without queries (2nd row)
	% with attribute based parsing 
	(81.8\% vs.\ 80.4\%). This is most likely due to: 
	(i)~data re-use enabled by shared sub-labels; and (ii)~the learnt queries 
	act as `experts' to identify discriminative events.
	
	\subsection{Feature Bank Ablations}\label{s:ft-bank-ablation}
	We benchmark our stochastically updated feature banks (\Cref{sec:ft-bank}) in 
	two ways: first, we evaluate the effect of increasing the  temporal context 
	during training, and second the effect of backpropagation through the feature 
	bank. We use Diving48-v2 for both.

	\paragraph{Effect of increasing training temporal context.}
	To evaluate the importance of dense and long temporal context during training for 
	fine-grained action recognition, we train the S3D visual encoder~\cite{s3d} on
	an increasing number of input frames $N = \{8, 32, 64, \text{all frames}\}$, 
	where `all frames' corresponds to LT-S3D, \ie training with our \sufb.
	% Note, `all frames' {\em require} our \sufb to densely-sample clips 
	% from videos at training time.
%	** not clear what this means. Is this LT-S3D or something else? **.
	At inference, full temporal support is used for all methods by averaging class
	probabilities from multiple clips spanning the entire video (no decoder). 
	To control for visual discontinuity between frames due to large input stride, we 
	sample the frames in two ways:
	(i)~\emph{consecutively} sample $N$ frames with a temporal stride of 2 starting at random temporal locations, and 
	(ii)~\emph{uniformly} sample $N$ frames with a span of the entire video, 
	where the temporal stride $s$ is proportional to the actual length $T$ of the video,
	\ie, $s = \lfloor \frac{T}{N}\rfloor$.
	% , using all frames is not feasible without feature banks).
	% Average-pooling over class probabilities obtained from multiple-clips 
	% is used for temporal aggregation (no decoder).
	% (similar to standard practice~\cite{i3d,s3d}).
	% Full results are presented in \sref{t:test-all-frames} due to space constraints. 
	\Cref{t:test-all-frames} summarizes the results. 
	Consecutive sampling performs better as uniform sampling implies varying 
	input stride which is not amenable to S3D's temporal convolutional filters 
	with fixed stride.
	% Although uniform sampling across the video 
	% gives a longer span, its results are worse due to the inconsistent stride. 
	More importantly, longer temporal context is better regardless of the frame 
	sampling strategy: 
	all frames: 80.5\% per-video accuracy; for $N = \{8, 32, 64\}$: ${<}75$\%.
	This demonstrates the critical role of our \sufb for training.
	
	\paragraph{Effect of updating bank features during training.}\label{s:exp-mbank}
	A key difference between our \sufb and previous methods for extended
	temporal support for videos, \eg, {\em Long-term feature banks}~\cite{featbank}
	and \cite{li2017temporal, miech2017learnable, tang2018non}, is that
	they use {\em frozen} features, while we continuously update them.
	We train with frozen/updated features, with/without the \TQN decoder on top
	of visual features, and summarize the results in \Cref{t:mbank}.
	For both with/without \TQN, updating the features improves the performance 
	substantially ($\approx +15\%$). Ablation study of the effect of number of features update on final performance can be found in \sref{s:supp-two-stage}.
			
	\begin{table}[]
		\vspace{-2mm}
		\renewcommand{\arraystretch}{1.2}
		\centering
		\resizebox{\linewidth}{!}{%
			\begin{tabular}{l|c|c|c|cc}
				\toprule
				\multicolumn{1}{c|}{\multirow{2}{*}{\textbf{Description}}} & \multirow{2}{*}{\textbf{\begin{tabular}[c]{@{}c@{}}Visual\\ Encoder\end{tabular}}} & \multirow{2}{*}{\textbf{Memory bank}}                                        & \multirow{2}{*}{\textbf{Decoder}}                   & \multicolumn{2}{c}{\textbf{Accuracy}}                        \\ \cline{5-6} 
				\multicolumn{1}{c|}{}                                      &                                                                                    &                                                                              &                                                     & \multicolumn{1}{c|}{\textbf{per-class}} & \textbf{per-video} \\ \hline
				train linear classifier on frozen encoder                  & \multirow{2}{*}{frozen}                                                            & \multirow{2}{*}{\begin{tabular}[c]{@{}c@{}}computed \\ offline\end{tabular}} & \begin{tabular}[c]{@{}c@{}}avg\\ pool\end{tabular} & \multicolumn{1}{c|}{57.1}               & 66.6               \\ \cline{1-1} \cline{4-6} 
				train TQN on frozen encoder                                &                                                                                    &                                                                              & TQN                                                 & \multicolumn{1}{c|}{60.9}               & 68.2               \\ \hline
				train linear classifier + encoder                          & \multirow{2}{*}{fine-tuned}                                                        & \multirow{2}{*}{\begin{tabular}[c]{@{}c@{}}updated\\ online\end{tabular}}    & \begin{tabular}[c]{@{}c@{}}avg\\ pool\end{tabular}  & \multicolumn{1}{c|}{72.3}               & 80.4               \\ \cline{1-1} \cline{4-6} 
				train TQN + encoder                                        &                                                                                    &                                                                              & TQN                                                 & \multicolumn{1}{c|}{74.5}               & 81.1               \\ \bottomrule
			\end{tabular}%
		}
		\vspace{1mm}
		\caption{{\bf Frozen vs.\  updated feature bank.} 
			To study the importance of updating the feature-bank during training, 
			we train with the \TQN decoder and without it (average
			pooling for temporal aggregation), on top of {\em frozen} or 
			{\em stochastically updated} visual features in the memory bank.
			For both the decoder settings, updating features improves performance
			drastically ($\approx +15\%$).
			Evaluation on the Diving48-v2 dataset; details in \Cref{s:ft-bank-ablation}.}
		\label{t:mbank}
	\end{table}

	\subsection{Comparison with State-of-the-art}\label{s:exp-sota}
	Finally, in \Cref{t:sota} we compare the 
	performance of  \TQN against SotA methods on Diving48-v2, 
	Gym99 and Gym288. For completeness, performance on the original noisy 
	Diving48-v1 is reported in \sref{s:noisy-diving}. 
	\TQN outperforms all methods on all the three benchmarks on both per-video and
	per-class measures, even when flow+RGB (two-stream) input is allowed for 
	other methods,  while only RGB is input to \TQN; a detailed table with breakdown
	for RGB and flow is included  in \sref{s:full-sota}.
	Compared to the ST-S3D baseline (S3D with short temporal context), having
	long-term context (LT-S3D) using our \sufb leads to drastic improvements: 
	$>$30\% (absolute) on Diving48-v2, and $>$10\% (absolute) on the Gym datasets.
	Adding \TQN decoder on top of LT-S3D leads to further improvements, notably
	on the `VT' (vaulting) subset of Gym99 (+5.8\%) which contains longer videos: $7{-}8$
	seconds compared to $1{-}2$ seconds in the `FX' (floor exercise) subset (+1.2\%).
	Note, the visual backbone of \TQN can be made stronger by replacing S3D with, 
	\eg, TSM, TSN, or GST-50. However, we adopt S3D in our 
	experiments as it achieves top performance while fitting within our limited 
	compute budget.
	
	\begin{table}[]
		\renewcommand{\arraystretch}{1.2}
		\resizebox{\linewidth}{!}{%
			\begin{tabular}{c|c|c|c|c}
				\hline
				\multicolumn{4}{c|}{\textbf{Training}}                                                                                                                                                              & \textbf{Test} (w/ all frames) \\ \hline
				temporal support                                                                   & \# frames ($N$)                                                                 & frame sampling & stride & per-video  acc                                                \\ \hline
				\multirow{6}{*}{\begin{tabular}[c]{@{}c@{}}fixed number\\  of frames\end{tabular}} & \multirow{2}{*}{8}                                                          & consecutive    & 2      & 58.8                                                           \\ \cline{3-5} 
				&                                                                             & uniform        & --      & 50.6                                                           \\ \cline{2-5} 
				& \multirow{2}{*}{32}                                                         & consecutive    & 2      & 72.9                                                           \\ \cline{3-5} 
				&                                                                             & uniform        & --      & 71.2                                                           \\ \cline{2-5} 
				& \multirow{2}{*}{64}                                                         & consecutive    & 2      & 74.2                                                           \\ \cline{3-5} 
				&                                                                             & uniform        & --      &                         70.0                                       \\ \hline
				\begin{tabular}[c]{@{}c@{}}full temporal \\ support\end{tabular}                   & \begin{tabular}[c]{@{}c@{}}proportional to\\  length of videos\end{tabular} & memory bank    & 2      & \textbf{80.4 }                                                          \\ \hline
		\end{tabular}}
		\vspace{1mm}
		\caption{\textbf{Impact of temporal support during training.}
		To analyze 
			the importance of temporal support for training and use of \SUFB,  we train S3D with an 
			increasing number of input frames $N$
			and find that longer temporal context consistently improves performance on Diving48.}
		\label{t:test-all-frames}
		\vspace{-1mm}
	\end{table}

	\begin{figure}[b]
		 \vspace{-3mm}
		\centering
		\begin{minipage}{.23\textwidth} 
			\centering
			\includegraphics[width=1.1\textwidth]{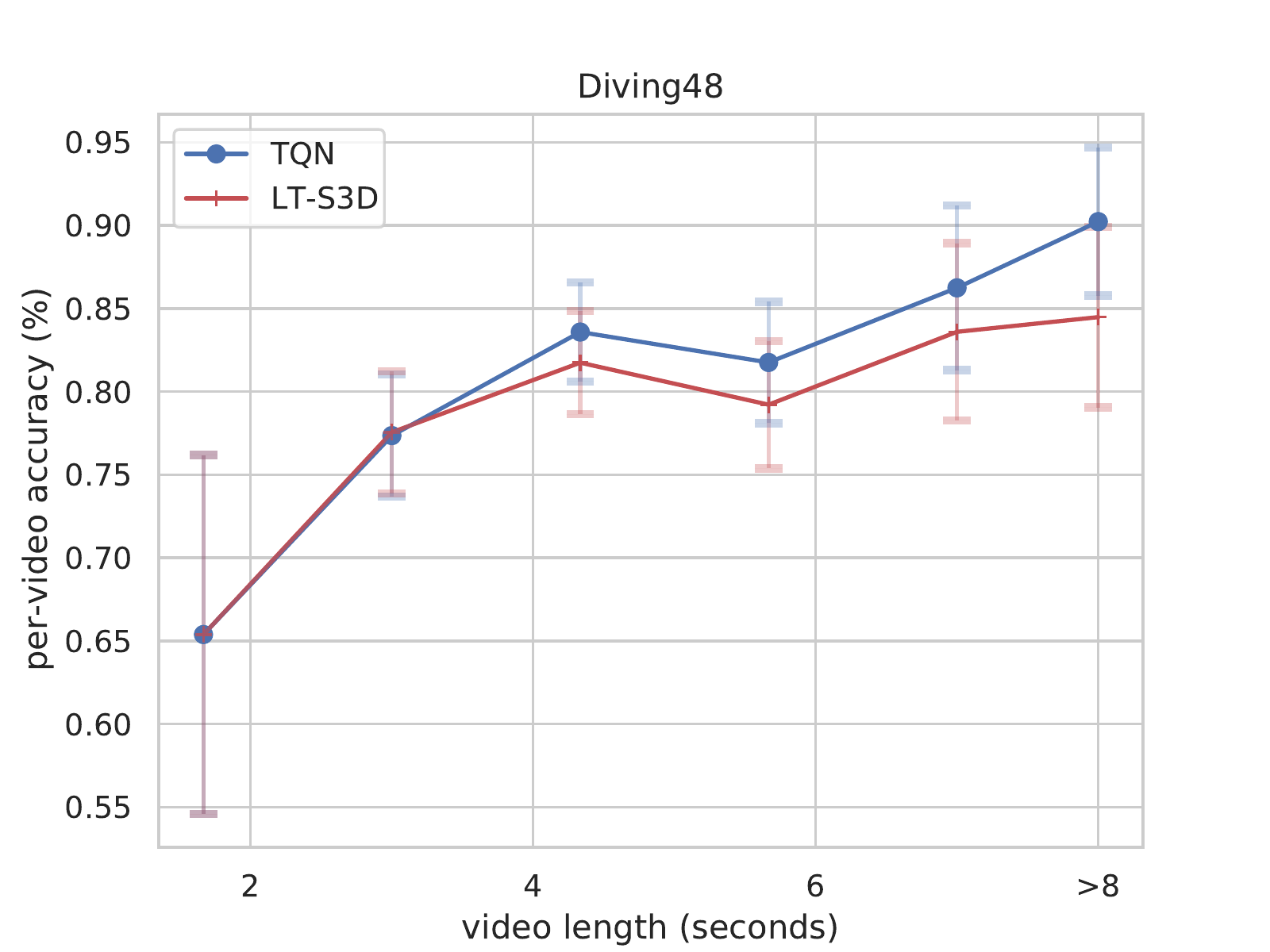}
		\end{minipage}
		\begin{minipage}{.23\textwidth}
			\centering
			\includegraphics[width=1.1\textwidth]{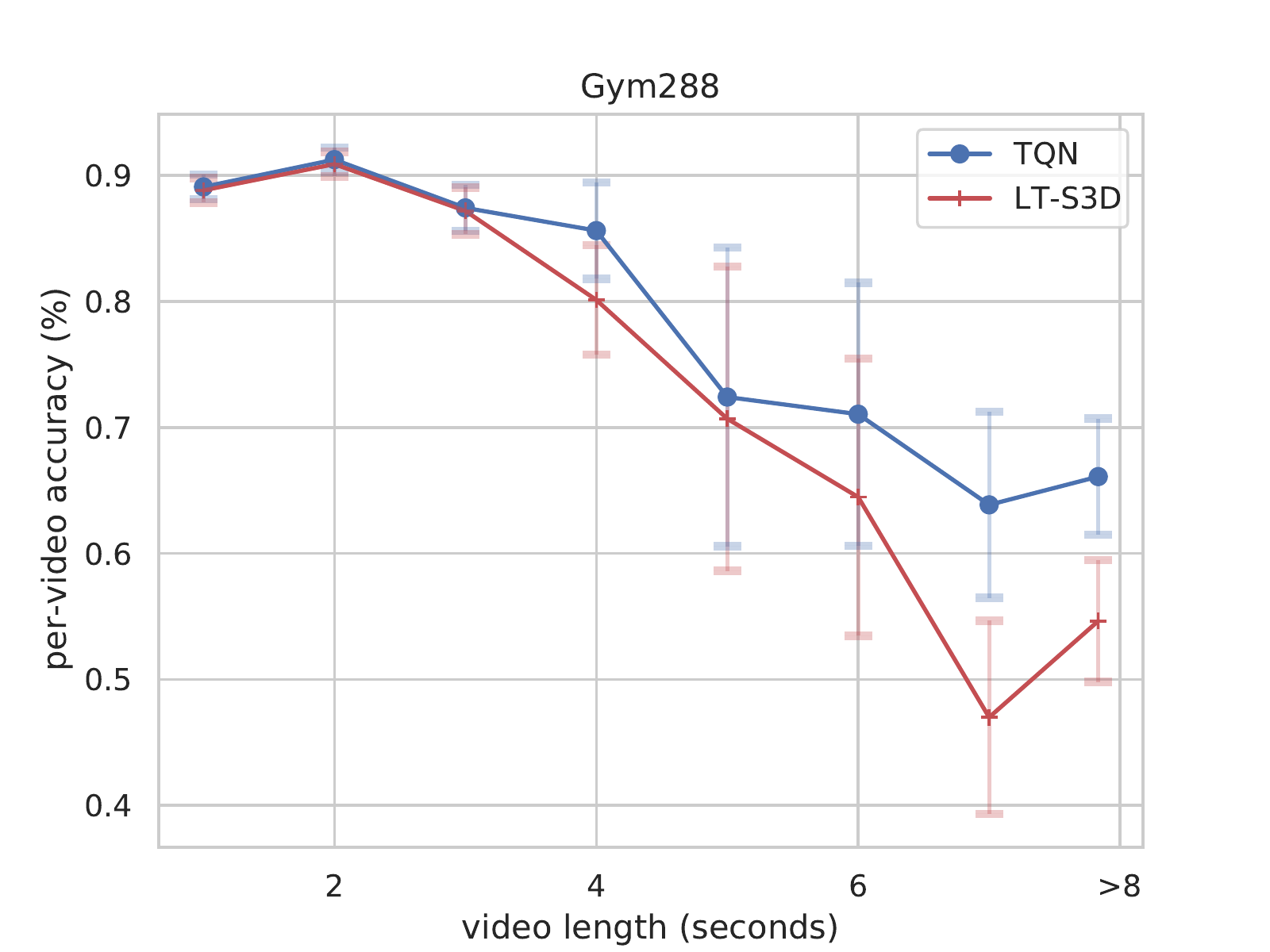}
		\end{minipage}
		\caption{{\bf Classification accuracy on videos of different length.} Mean values plotted with 95\% confidence interval.}
		\label{f:acc_vs_leng}
	\end{figure}

\subsection{Performance on Videos of Different Length}
In \Cref{f:acc_vs_leng} we plot classification accuracy of  \TQN and 
the baseline long-term S3D as a function of video length.
On videos shorter than 5 seconds, LT-S3D performs similar to \TQN
as max-pooling suffices to pick out relevant information in short videos. 
However, \TQN's attention based classification outperforms simple pooling for 
longer videos.
% The improvement brought by TQN is larger with increasing length of videos due to its advantage in modelling long-term temporal features.

\subsection{Transfer learning with TQN}\label{s:exp-transfer}
To investigate the transferability of \TQN across domains which differ 
visually as well as in their query-attributes, we fine-tune the model 
pre-trained on Gym288 for Diving48.
The query vectors $\bq$ and the response classifiers ${\bf\Psi}$ are tied to 
dataset specific query-attributes. Hence, to fine-tune on a new dataset,
we initialize these randomly and retain the initialization from pre-training 
for other \TQN and visual backbone parameters. We compare this to the random 
initialization baseline in \Cref{t:transfer}. We note that fine-tuning gives 
better accuracy and trains substantially faster as compared to training 
from scratch. This is likely because the transformer has 
learnt (and retained) how to match query vectors to temporal events, and encode 
the event representation for response classifiers. It is thus able to benefit 
from the additional training data despite the domain shift.

\begin{table}[]
	\centering
	\resizebox*{.70\linewidth}{!}{%
		\begin{tabular}{@{}lcc@{}}
			\hline
			\multicolumn{1}{c}{\textbf{Pre-training}} & \textbf{Epochs} & \multicolumn{1}{c}{\textbf{Diving48-v2 top-1}} \\ \hline
			None (random init.)                        & 80                              & 81.8                                       \\
			Gym288                                    & {\bf 25}                              & {\bf 83.3}                                       \\ 
			\hline
	\end{tabular}}
	\vspace{1mm}
	\caption{\textbf{Transferring TQN.} Per-video performance on Diving48-v2 for a \TQN  model pre-trained on Gym288.}
	\label{t:transfer}
	\vspace{-2mm}
\end{table}

\subsection{Extension to Multi-label Action Recognition}
We apply \TQN to the Charades dataset~\cite{charades} and achieve comparable results with the SotA models. Charades labels multiple actions in one video as different classes with 
precise temporal annotations, as opposed to classification in FineGym and 
Diving48 where a sequence of combined actions are labelled as one class 
without localization.  Please refer to \sref{s:charades}.

% In Finegym and Diving48, a sequence of combined actions are labelled as a single 
% class without any temporal annotations. Therefore, manual query-attributes 
% factorization has to be done for TQN. However, in some other popular 
% multi-label action recognition datasets~\cite{charades,ava} actions in one 
% videos are labelled separately with precise temporal localization, where the 
% action classes can be considered as natural queries. We conduct experiments to 
% show that \TQN can be easily adapted to this task and achieve SotA 
% performance on Charades~\cite{charades}. Please refer to \sref{s:charades}.
\vspace{-2mm}
\section{Conclusion}
	We have developed a new video parsing model, the Temporal Query Network (\TQN), which learns to answer fine-grained questions about event types and their attributes in untrimmed videos.
	%  given only weak supervision, \ie, no explicit 
	% temporal position or duration information.
	\TQN furthers state-of-the-art in fine-grained video categorization on three datasets, and in addition provides temporal localization and alignment of semantically consistent events. The {\em query-response} mechanism employed in \TQN enables efficient data use through sharing training videos across common 
	sub-labels and outperforms alternative strategies for exploiting textual descriptions. The mechanism is more generally applicable to problems which require spotting entities with varying spans in dense data streams. Our training method with stochastically updated feature banks enables such applications without imposing heavy requirements for expensive large-scale training infrastructure.
% I don't think we need to disclose our future work here
%	Future work may include automating the generation of query-attributes set and alleviate the need to re-initialize them when transferring to a different dataset.

% Pros and Cons
%	\textbf{Pros:} (i)~allows (soft-) detection based classification in untrimmed videos without temporal supervision (line 97--105).
%	(ii)~efficient data use through sharing training videos across common 
%	sub-labels (line 557).
%	(iii)~outperforms alternative strategies for exploiting textual descriptions of
%	the categories (sec.~5.1) and achieves top results (table~3).
%	%
%	\textbf{Cons:} The query-attributes are (i)~pre-determined and fixed once 
%	the model is trained, hence, need re-initialization for transfer (line \ft above),
%	and (ii)~currently obtained manually, however, can be automated
%	
\small\noindent\textbf{Acknowledgements.} This research is funded by a Google-DeepMind
Graduate Scholarship, a Royal Society Research Professorship, and the EPSRC Programme Grants  Seebibyte EP/M013774/1 and  VisualAI EP/T028572/1. 
We thank Weidi Xie, Tengda Han and the reviewers for helpful insights.

	{\small
		\bibliographystyle{ieee_fullname}
		\bibliography{bib/shortstrings,bib/vgg_local,bib/vgg_other,bib/refs}}
	
	\ifarxiv   %% add supplementary as appendix for arxiv:
	\onecolumn
	\hypersetup{%
		colorlinks = true,
		linkcolor  = red
	}

	\appendix
	\appendixpage
	%\hypersetup{colorlinks = true, linkcolor  = black}

%\startcontents[sections]
%\printcontents[sections]{ }{1}{}
%\clearpage

%\hypersetup{colorlinks = true, linkcolor  = red }

\section{Visualization of Action Attention}
In \Cref{fig:supp-loc-viz_dive,fig:supp-loc-viz_gym} below, we show some samples with the predicted temporal attention  by queries. \Cref{fig:supp-loc-viz_dive} shows some visualizations from Diving48, it can be seen that the attention for \textit{`take off'} is  usually located at the beginning of the sequence, while the ones for the queries \textit{`turn'} and \textit{'twist'} span over multiple central clips. \Cref{fig:supp-loc-viz_gym} shows examples taken from FineGym, despite the variance between videos, the model always responds to \textit{`pose'} and \textit{`circle' } at the begining of the action, while to \textit{`turn' }and \textit{`stand'} towards the end of the action.

Please refer to our project page\footnote{\url{https://www.robots.ox.ac.uk/~vgg/research/tqn/}} for visualizations using motion clips (GIFs).

\begin{figure}[h]
	\centering
	\includegraphics[height=14cm,width=15.0cm]{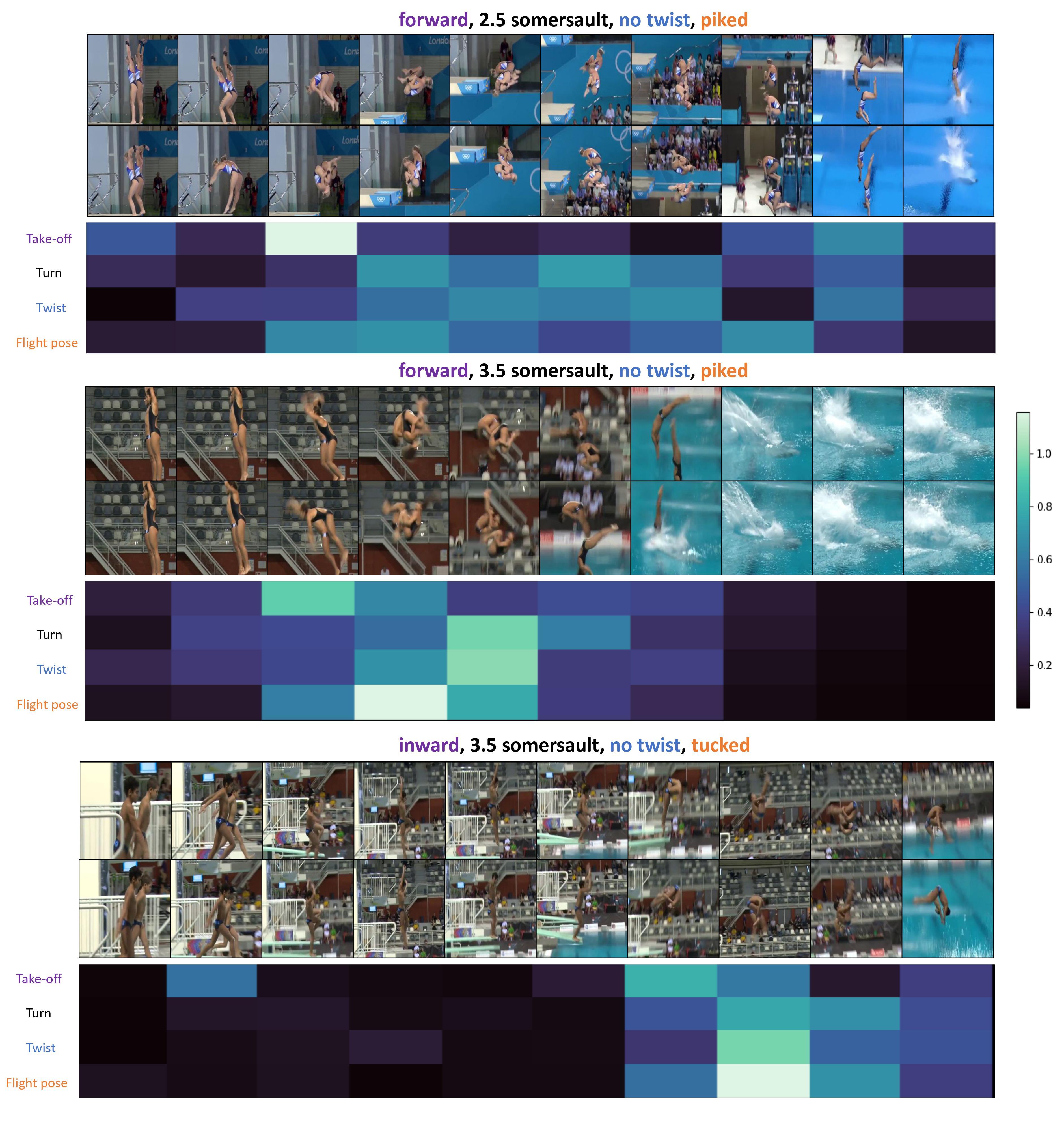}
	\captionof{figure}{\textbf{Visualization of temporal attention from four queries in Diving48.}  For every video sample, we show the attention scores over 10 clips from 4 queries by a 4x10 heatmap. Every clip contains 8 frames originally, here we show two from each (one above the other) for visualization.}
	\label{fig:supp-loc-viz_dive}
\end{figure}

\begin{figure}[h]
	\centering
	\includegraphics[height=19.5cm,width=19cm]{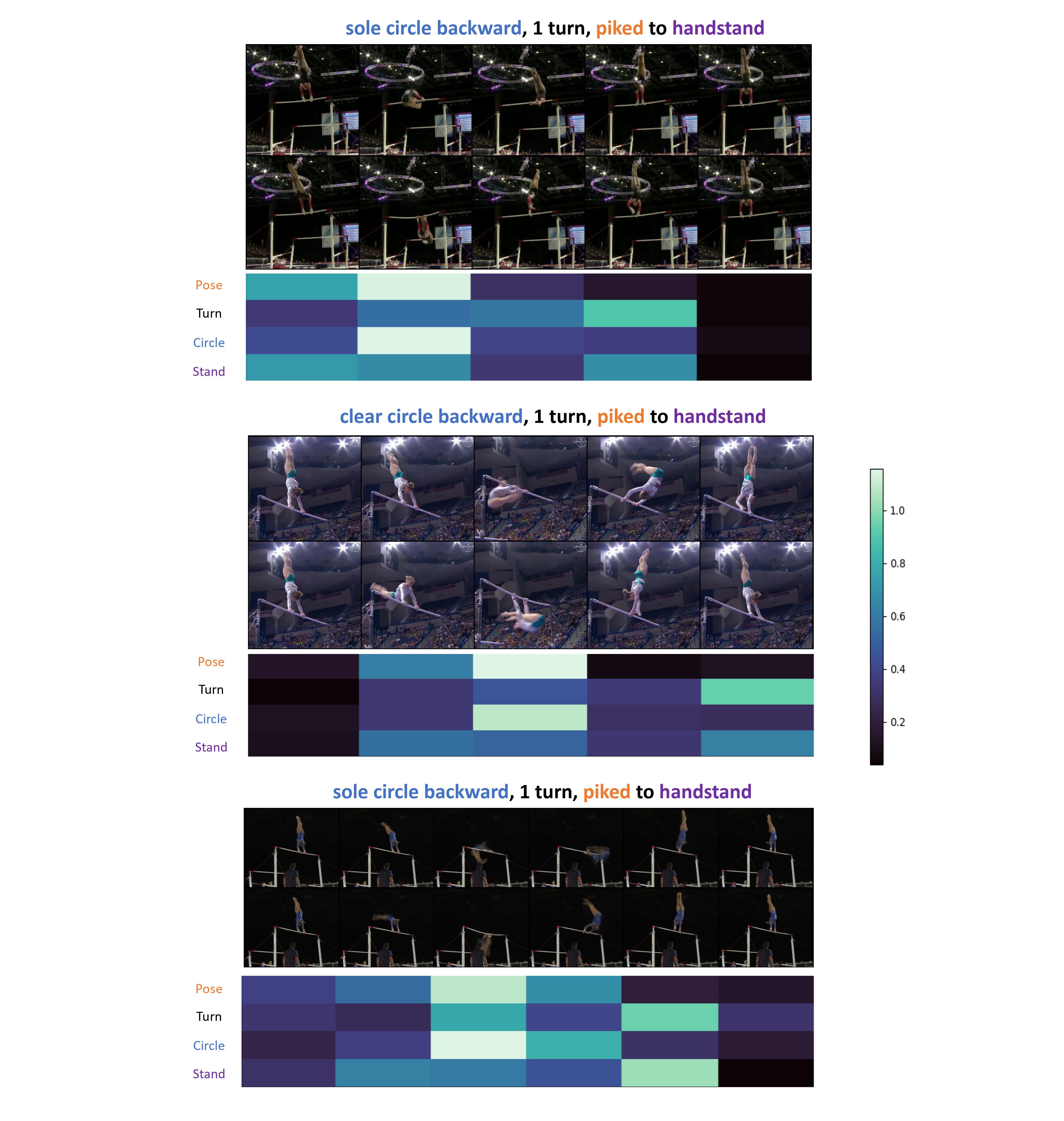}
	\captionof{figure}{\textbf{Visualization of temporal attention from four queries in Gym99.}  For every video sample, we show the attention scores over 5 clips from 4 queries by a 4x5 heatmap. Every clip contains 8 frames originally, here we show two from each (one above the other) for visualization.}
	\label{fig:supp-loc-viz_gym}
\end{figure}

\clearpage

\section{Full SotA Comparison}\label{s:full-sota}
% \ag{As discussed in email, we should preferably report flow numbers for us (and perhaps also for ST-S3D, LT-S3D, and GST-50).}

In FineGym~\cite{shao2020finegym}, SotA baselines like TSN~\cite{tsn}, TSM~\cite{tsm} and TRN~\cite{trn} are trained using both RGB and flow. Due to space limitations, in Table 3 in the main paper we show only
the best two-stream results from these model,  without breakdown for RGB and flow.  In \Cref{t:sota-supp} the detailed breakdown is shown: although the performance from two-streams is better than using only one modality for previous work, our model achieves superior results to all of these using only RGB frames.

\begin{table*}[t]
	\resizebox{\textwidth}{!}{%
		\begin{tabular}{cccc|cccc|cc|cc}
			\hline
			\multirow{3}{*}{\textbf{Network}} & \multirow{3}{*}{\textbf{Pretrained dataset}} & \multirow{3}{*}{\textbf{Modality}} & \multirow{3}{*}{\textbf{\# frames in training}} & \multicolumn{4}{c|}{\textbf{Gym99}}                         & \multicolumn{2}{c|}{\textbf{Gym288}}                    & \multicolumn{2}{c}{\textbf{Diving48}}                   \\ \cline{5-12} 
			&                                              &                                    &                                                 & \multirow{2}{*}{per-class} & \multicolumn{3}{c|}{per-video} & \multirow{2}{*}{per-class} & \multirow{2}{*}{per-video} & \multirow{2}{*}{per-class} & \multirow{2}{*}{per-video} \\ \cline{6-8}
			&                                              &                                    &                                                 &                            & subset VT  & subset FX & total &                            &                            &                            &                            \\ \hline
			I3D~\cite{i3d}                               & K400                                         & RGB                                & 8                                               & 64.4                       & 47.8       & 60.2      & 75.6  & 28.2                       & 66.7                       & 33.2                       & 48.3                       \\
			TSN~\cite{tsn}                                & ImageNet                                     & RGB                                & 3                                               & 68.7                       & 46.6       & 72        & 74.8  & 26.5                       & 68.3                       & 26.9                       & 41.3                       \\
			& ImageNet                                     & flow                               & 3                                               & 77.2                       & 42.6       & 81.2      & 84.7  & 38.7                       & 78.3                       & 28.1                       & 38.8                       \\
			& ImageNet                                     & two-stream                         & 3                                               & 79.8                       & 47.5       & 84.6      & 86    & 37.6                       & 79.9                       & 34.8                       & 52.5                       \\
			TSM~\cite{tsm}                               & ImageNet                                     & RGB                                & 3                                               & 70.6                       & 42.2       & 68.8      & 80.4  & 34.8                       & 73.5                       & 25.3                       & 38.0                       \\
			& ImageNet                                     & flow                               & 3                                               & 80.3                       & 42.4       & 81.9      & 87.1  & 46.8                       & 81.6                       & 25.3                       & 39.5                       \\
			& ImageNet                                     & two-stream                         & 3                                               & 81.2                       & 44.8       & 84.9      & 88.4  & 46.5                       & 83.1                       & 32.7                       & 51.1                       \\
			TRNms~\cite{trn}                             & ImageNet                                     & RGB                                & 3                                               & 68.8                       & 46.6       & 73.4      & 79.5  & 32.0                       & 73.1                       & 43.8                       & 56.8                       \\
			& ImageNet                                     & flow                               & 3                                               & 77.6                       & 43.9       & 81.1      & 85.5  & 43.4                       & 79.7                       & 46.4                       & 54.6                       \\
			& ImageNet                                     & two-stream                         & 3                                               & 80.2                       & 47.3       & 84.9      & 87.8  & 43.3                       & 82.0                       & 54.4                       & 66.0                       \\
			GST-50~\cite{gst}                               & ImageNet                                     & RGB                                & 8                                               & 84.6                       & 53.6       & 84.9      & 89.5  & 46.9                       & 83.8                       & 69.5                       & 78.9                       \\
			ST-S3D~\cite{s3d}                               & K400                                         & RGB                                & 8                                               & 72.9                       & 45.3       & 82.8      & 81.5  & 42.4                       & 75.8                       & 36.3                       & 50.6                       \\ \hline
			LT-S3D                              & K400                                         & RGB                                & dense                                           & 88.9                       & 69.1       & 90.4      & 92.5  & 57.9                       & 86.3                       & 72.3                       & 80.4                       \\
			TQN                               & K400                                         & RGB                                & dense                                           & \textbf{90.6      }                 & \textbf{74.9   }    & \textbf{91.6  }    & \textbf{93.8}  & \textbf{61.9  }                     & \textbf{89.6  }                     & \textbf{74.5 }                      & \textbf{81.8  }                     \\ \hline
		\end{tabular}%
	}
	\vspace{3mm}
	\caption{\textbf{Full Comparison to SotA results on Gym99, Gym288, Diving48. with breakdown for RGB and flow.}}
	\label{t:sota-supp}
\end{table*}

\section{Extension to Multi-label Action Recognition}\label{s:charades}
We apply \TQN to the Charades dataset~\cite{charades} and evaluate its performance on multi-label action recognition. Results are shown in \Cref{t:charades-loc}.

Charades contains 9.8k training and 1.8k test crowd-sourced videos
(avg. length: 30 secs) of scripted indoor activities, with each video containing one 
or more actions (avg. duration = 12 secs/action) out of 157 total different action classes.
As opposed to classification in FineGym and 
Diving48 where a sequence of combined actions are labelled as one class 
without localization, Charades labels multiple actions in one video as different classes with 
precise temporal annotations. 

Since the action classes in Charades do not form any natural clustering, we treat each class as an individual query and train the TQN to make binary classification. More specifically, the 157 learnable queries corresponds to 157 classes, and their output predict whether these actions exist or not. For training, we follow two-stage training pipeline in ~\cite{featbank}. In the first stage, we train the model on clips of ~3.2 seconds to make clip-level prediction. In the second stage, we add in the \SUFB to train the model for video-level prediction with dense features sampled from the entire video sequence.

During inference, we follow~\cite{featbank} and use 3-crop test to evaluate two models: 1) The model trained without \SUFB in the 1st stage. This model makes clip-level predictions which are aggregated by maxpooling to output video-level predictions. 2) The model trained with \SUFB using two-stage training. 

Results in  \Cref{t:charades-loc} show that when trained on short clips, the TQN architecture is better than most of the other architectures. When the \SUFB is used in training, the performance is further improved, leading to a better mAP compared to others using the same visual backbone pretrained on Kinetics-400. 

\renewcommand{\arraystretch}{1.3}
\begin{table}[]\tiny
	\centering
	\resizebox{0.5\textwidth}{!}{%
		\begin{tabular}{ccccc}
			\hline
			& \textbf{Backbone}   & \textbf{Feature bank} &   \textbf{Pretrain} &\textbf{Train/val} \\ \hline
			Asyn-TF~\cite{asyn} & VGG16      & \xmark     & ImageNet & 22.4      \\
			TRNms~\cite{trn}   & Inception  & \xmark     & ImageNet      & 25.2      \\
			I3D~\cite{i3d} (from ~\cite{nonlocal})      & R101-I3D   & \xmark     & K400      & 35.5      \\\hline
			I3D-NL~\cite{nonlocal} (from ~\cite{featbank})  & R50-I3D-NL & \xmark     & K400      & 37.5      \\
			STRG~\cite{STRG}    & R50-I3D-NL & \xmark     & K400      & 37.5      \\
			
			STO~\cite{featbank}     & R50-I3D-NL &     \xmark     & K400      & 39.6      \\
			LFB NL~\cite{featbank}  & R50-I3D-NL & \cmark     & K400      & 40.3      \\ 
			TQN  & R50-I3D-NL & \xmark     & K400      & 40.7      \\
			TQN  & R50-I3D-NL & \cmark     & K400      &\textbf{ 41.4 }     \\ \hline
			STO~\cite{featbank} & R101-I3D-NL &\xmark     & K400      & 41.0     \\
			SlowFast NL~\cite{slowfast}  & R101-I3D-NL &\xmark     & K400      & 42.5      \\
			LFB NL~\cite{featbank} & R101-I3D-NL &\cmark     & K400      & 42.5     \\
			TQN  & R101-I3D-NL & \xmark     & K400      &42.1     \\
			TQN  & R101-I3D-NL & \cmark     & K400      &\textbf{42.9 }    \\
			\hline
			\color{gray}SlowFast NL~\cite{slowfast}   &\color{gray} R101-I3D-NL &\color{gray}\xmark     & \color{gray}K600      & \color{gray} 45.2     \\
			\hline
		\end{tabular}%
	}
	\caption{\textbf{Comparison to state-of-the-art on Charades when using temporal localization.}}
	\label{t:charades-loc}
\end{table}

\section{Implementation Details}\label{s:supp-impl}

\paragraph{Decoder architecture.}%
The decoder has the same architecture as a  basic Transformer decoder~\cite{vaswani2017attention}. The decoder takes features from the 
visual encoder for keys and values, and the learnt embeddings as queries. In every attention layer, the decoder 
attends to keys and aggregate the values to update the queries. The updated queries serve as new input queries to the next layer which interact with visual features in the same way. Linear classifiers are applied to the output features from the decoder for attribute prediction.

In our experiments on FineGym and Diving48, we use four attention layers with four heads with 1024 feed-forward dimension, with dropout rate equal to 0.1 in the decoder and 0.5 for output features.
In our experiments on Charades,  we use two attention layers with two heads with 1024 feed-forward dimension, with dropout rate equal to 0.1 in the decoder and 0.3 for output features.
We omit positional encoding in all the experiments as we find it leads to overfitting in this task.

\paragraph{Data augmentation.}%
The input images are non-overlapping contiguous video clips of 8 frames with temporal stride of 1 in FineGym,  8 frames with temporal stride of 2 in Diving48 and 32 frames with  temporal stride of 4 in Charades. In all the training, we apply color jittering (brightness factor $\in$[0.6,1.4], contrast factor $\in$ [0.3,1.7], saturation factor $\in$ [0.3,1.7], hue $\in$ [-0.25,0.25]) and horizontal flips (with probability 50\%) on images of size 224 x 224 pixel.

In training of FineGym and Diving, we apply spatial and temporal augmentation. In the spatial dimension, we randomly crop the frames to squares of 60\%-100\% of the original height, and resize them to $224{\times}224$ pixels. In the temporal dimension, we randomly remove 2\% of frames when sampling as augmentation. In training of Charades, we follow the random spatial crop in \cite{featbank} without further temporal augmentation. 

\paragraph{Inference.}%
When testing \TQN, all the features from one video are computed online without using the feature bank. 

When TQN is tested on FineGym and Diving, only center-crops are used. For every sample, we randomly remove 2\% of frames three times to obtain three sequences, and average the class probabilities from them to obtain final predictions. For other baseline models which are trained on a fixed number of frames but tested with all frames (\eg, ablation studies in Section 5.2 in the main paper), we use multiple clips in inference. We randomly/consecutively sample multiple clips (with the same number of frames per clip as in training) so that they cover the full video sequence, and average the class probabilities from all the clips to obtain final predictions.

On Charades, we follow ~\cite{nonlocal} and use 3 spatial crops for each clip in inference without temporal jittering. The predictions from 3 crops are aggregated by maxpooling the class probabilities.

\paragraph{Learning rate schedule.}%
For FineGym and Diving48, no temporal localization information is used in the training. Our training pipeline contains two stages: (1) end-to-end training on short videos as warm-up to obtain a good initialization. (2) Training with a stochastically updated memory feature bank on the longer videos in a bootstrapping way. 

we use Adam optimizer with  base learning rate 0.001 in the encoder and 0.0001 in the decoder, and the weight decay is 1e-5. 
For Charades, we use SGD optimizer with base learning rate 0.02  with momentum 0.9, weight decay is set to 1.25e-5. Detailed schedule of training is shown in \Cref{tab:train_stage}.

When temporal localization is used in training (Charades), we first use the localization to train the model to do clip-level actions as in \cite{featbank} for 45 epochs with learning rate 0.01 decreased by 10 at epoch 35, and then use stochastically updated memory feature bank to train the model to make video-level actions using learning rate 0.01 for 16 epochs,decreased by 10 at epoch 6.

\begin{table}[]
	\renewcommand{\arraystretch}{1.2}
	\centering
	\resizebox{0.6\textwidth}{!}{%
		\begin{tabular}{c|ccc|ccc}
			\hline
			\multirow{2}{*}{\textbf{Dataset}} & \multicolumn{3}{c|}{\textbf{1st stage}}                                                                                                           & \multicolumn{3}{c}{\textbf{2nd stage}}                                                                                                                                                                                                                                               \\ \cline{2-7} 
			& \# epochs & \begin{tabular}[c]{@{}c@{}}sequence shorter \\ than N frames\end{tabular} & \begin{tabular}[c]{@{}c@{}}lr \\ multiplier\end{tabular} & \# epochs & \begin{tabular}[c]{@{}c@{}}sequence shorter \\ than N frames\end{tabular} & \begin{tabular}[c]{@{}c@{}}lr \\ multiplier\end{tabular}                                  \\ \hline
			FineGym                           & 50        & 48                                                                        & 1                                                        & 35        & max (all videos)                                                          & \multirow{2}{*}{\begin{tabular}[c]{@{}c@{}}0.1 @epoch 10 ,\\ 0.01 @ epoch 20 \end{tabular}}                                   \\ \cline{1-6}  
			Diving48                          & 50        & 128                                                                       & 1                                                        & 35        & max (all videos)                                                          &                                                                                                              \\ \cline{1-7} 
			Charades                          & 50        & 128                                                                       & 0.1 @epoch 35 , 0.01 @ epoch 45                                                       & 35        & max (all videos)                                                          &                   0.1 @ epoch 6                                                                                              \\ \cline{1-7}                                     
		\end{tabular}%
	}
	\caption{\textbf{Details of multi-stage training with stochastic feature bank on different datasets.}}
	\label{tab:train_stage}
\end{table}
%We train the model for 50 iterations in the first stage, the learning rate (LR) is set to constant values of 1e-3 in the visual encoder and 1e-4 in the query decoder. 
%In the second stage the model  is trained with the memory bank on the entire dataset for 30 epochs, the learning rates drops every 10 epochs. More specifically, in the $3{\times}10$ epochs, the LR is set to 1e-3, 1e-4, 1e-5 in the encoder, and 1e-4, 1e-4, 1e-5 in \TQN decoder respectively. 

\paragraph{Feature bank.}%
Before we use the feature bank, we first train the model end-to-end on videos shorter than $N$ clips (detailed in \cref{tab:train_stage}). We use this model to initialize the feature bank: feature vectors from all training videos are extracted and cached. The feature bank storage takes 452Mb for Charades, 725Mb for Diving48, 871Mb for Gym99 , and 1.5G for Gym288 (40k training videos, 6 clips in each video on average). During training with the stochastically updated feature bank, at each iteration, features from $n_{\text{online}}$ randomly selected contiguous clips are computed online and cached in the bank; $n_{\text{online}}$ = 10 in Diving48, $n_{\text{online}}$ = 6 in FineGym, and $n_{\text{online}}$ = 3 in Charades. 
For inference, all the features from one video are computed online without using a feature bank.

\section{Ablation on Two-Stage Training} \label{s:supp-two-stage}
We adapt the two-stage training to obtain good performance on long videos.  In the warm-up stage where the model is trained end to end efficiently on short videos with the number of clips up to $k_{\text{first}}$, where 1 clip spans over 16 frames, with temporal downsampling rate equal to 2. And in the second stage we train the model with the feature bank on the whole dataset containing both short and long videos. In every iteration, we compute features from $n_{\text{online}}$ clips online and update them in the feature bank. Here we ablate how the choice of$k_{\text{first}}$ and $n_{\text{online}}$ affects the final performance.  

We use $k_{\text{first}}$ = 8 and $n_{\text{online}}$ = 10 in all the experiments presented in the main paper. In the following experiments, we use change one of them to different values and keep the other fixed, and use the Diving48 dataset for evaluation.
% Results are shown in \Cref{tab:1st_stage_ablation} and \Cref{tab:2nd_stage_ablation}. 

In \Cref{tab:1st_stage_ablation}, we have $k_{\text{first}}$ $\in \{0,4,6,8\}$ and keep $n_{\text{online}}$ constant as 10.  When  $k_{\text{first}}$ there is no warm-up training, we use the model pre-trained on Kinetics400~\cite{kinetics} as initialization and train with the feature bank directly, the per class and per videos accuracy is 72.3\% and 79.2\% respectively. When warm-up training  is done on videos shorter than 4 clips, it actually worsens the performance, as such videos only account for 5\% of the whole dataset, resulting in overfitting. 
When $k_{\text{first}}$ is further increased to 8, it gives improvement on both per-video and per-class accuracy by 2\%.

In \Cref{tab:2nd_stage_ablation}, we keep $k_{\text{first}}$ fixed as 8 and increase  $n_{\text{online}}$ from 0 to 10 (0 meaning no 2nd stage training using memory banks).  Increasing the size of  $n_{\text{online}}$ also gives better performance. Both per-class accuracy and per-video accuracy is improved by 2\% when $n_{\text{online}}$ is doubled from 4 to 8.

Therefore, both two stages are necessary to obtain good results. In the first stage the model is trained efficiently on short videos. It prevents the model from overfitting to uninformative clips in the second stage, especially when  a small number of clips are updated online. The second stage makes it possible to train the model on long sequences which are impossible to fit into first stage. We use two RTX 6000 GPUs (total 48 GB memory) in our main experiments, if more GPU memory is available, $k_{\text{first}}$ and $n_{\text{online}}$ could be increased to even higher numbers for better performance.

\renewcommand{\arraystretch}{1.5}
\setlength{\tabcolsep}{2pt}

\begin{minipage}{\textwidth}
	\captionsetup{font=footnotesize}
	\begin{minipage}[!htb]{0.45\textwidth}
		\scriptsize
		\centering
		
		\begin{tabular}{c|c|c}
			\hline
			\multirow{2}{*}{\textbf{\begin{tabular}[c]{@{}c@{}}Use videos up to $k_{\text{first}}$ clips in\\ warm-up training\end{tabular}}} & \multicolumn{2}{c}{\textbf{Accuracy}} \\ \cline{2-3} 
			& per class          & per video         \\ \hline
			
		$k_{\text{first}}$ = 8                                                                                                                 & 74.5               & 81.8              \\ \hline
		$k_{\text{first}}$= 6                                                                                                                 &   69.4    &    79.7   \\ \hline
		$k_{\text{first}}$= 4                                                                                                                &     67.5             &        78.9          \\ \hline
		$k_{\text{first}}$= 0                                                                                                          &  72.3       &         79.2       \\ \hline
		\end{tabular}%
		\captionof{table}{\textbf{Ablation of max length of videos used in warm-up training (the 1st stage in 2-stage training).} \label{tab:1st_stage_ablation}	}

	\end{minipage}
	\hfill
	\begin{minipage}[!htb]{.45\textwidth}
		\scriptsize
		\centering
		\begin{tabular}{c|c|c}
			\hline
			\multirow{2}{*}{\textbf{\begin{tabular}[c]{@{}c@{}}$n_{\text{online}}$ clips updated \\ in feature bank\end{tabular}}} & \multicolumn{2}{c}{\textbf{Accuracy}} \\ \cline{2-3} 
			& per class          & per video         \\ \hline
			$n_{\text{online}}$ = 10                                                                                                                            & 74.5               & 81.8              \\ \hline
			$n_{\text{online}}$ = 8                                                                                                                             &      71.1              &   80.3                \\ \hline
			$n_{\text{online}}$ = 6                                                                                                                             & 70.4            & 79.7             \\ \hline
			$n_{\text{online}}$ = 4                                                                                                                         &     69.8        &   78.3        \\ \hline
			$n_{\text{online}}$ = 0                                                                                                                         &     64.4        &   75.8        \\ \hline
		\end{tabular}%
		\captionof{table}{\textbf{Ablation of number of clips updated in the feature bank (the 2nd stage in 2-stage training).}\label{tab:2nd_stage_ablation}}
	\end{minipage}

\end{minipage}

\section{Feature Bank vs.\ End-to-end Training}

% \begin{table}[h]
\setlength{\columnsep}{10pt}
\begin{wraptable}[14]{r}{7.0cm}
	\vspace{-3mm}
	\centering
	\resizebox{5.2cm}{!}{%
		\begin{tabular}{c|c|c|c}
			\hline
			\multirow{2}{*}{\textbf{Method}}       & \multirow{2}{*}{\textbf{\begin{tabular}[c]{@{}c@{}}\# clips updated\\ in feature bank\end{tabular}}} & \multicolumn{2}{c}{\textbf{Accuracy}} \\ \cline{3-4} 
			&                                                                                                     & per-class     & per-video     \\ \hline
			\multirow{3}{*}{\textbf{Feature bank}} & 4                                                                                                &     70.2          &        79.3       \\ \cline{2-4} 
			& 8                                 &        71.7    &          81.3     \\ \cline{2-4} 
			& 12                                                                               &      73.0         &         81.4      \\ \hline
			End-to-End                             & --                                                                                                   &      73.9         &         81.6      \\ \hline
		\end{tabular}%
	}
	\caption{\textbf{Performance gap between training with the feature bank and end-to-end training.}  Comparison is done on videos shorter than 256 frames so that end-to-end training is possible with our memory capacity.}
	\vspace{8mm}
	\label{tab:mbank}
\end{wraptable}
% \end{table} 

To benchmark how closely the stochastically updated feature bank can approximate the ideal end-to-end training, we investigate the performance gap between the two settings. Comparisons are done on videos shorter than 16 clips (256 frames) in Diving48 so that the full sequence can fit into memory of four RTX 6000 (96GB total) (we use two extra GPUs here for benchmarking feature banks; all other experiments use two GPUs due to limited budget). The training pipeline is the same as the one in the main paper except the 1st stage training is done on videos shorter than 8 clips instead of 10 clips here (as the max length of videos are now smaller).

The first three rows in \Cref{tab:mbank} show the results from using the stochastically updated feature bank, with different update step sizes in one iteration. The last row shows the results from end-to-end training with densely sampled frames from the whole video. The gap between updating 12 clips (75\% of max length) and end-to-end training is less than 1\%. Hence, with good initialization and large update step, the performance from using the feature bank can be close to the one from end-to-end training. 

In our experiments in Diving48 in the main paper, our update step size is 10 clips (160 frames, 50\% of the max length). 
The accuracy (81.8\% per video) might be further improved with more GPU memory and larger step size.

\section{Text Supervision}\label{s:text-sup}
In Section 5.1  in the main paper, we explore different ways of leveraging multi-attribute labels, and compare them to standard multi-class classification on Diving48, the results are reproduced in \Cref{tab:texts}. Here we describe the details of all the methods and their architectures, and analyze their performance on fine-grained action recognition.
For fair comparison, we use S3D as the visual encoder for all the architectures below. When a Transformer encoder or decoder is used in the method, we construct them with 4 attention layers and 4 attention heads so that their number of parameters is similar or the same as our \TQN decoder.

\begin{table}[h]
	\centering
	\renewcommand{\arraystretch}{1.2}
	\resizebox{0.55\linewidth}{!}{
		\begin{tabular}{c|c|c|c|c|c|c}
			\hline
			\multirow{2}{*}{\bf Backbone} & \multirow{2}{*}{\bf Encoder}        & \multirow{2}{*}{\begin{tabular}[c]{@{}c@{}}{\bf Decoder}\\ (Aggregation)\end{tabular}} & \multirow{2}{*}{\bf Classification}                                                        & \multirow{2}{*}{\bf Label}                                                        & \multicolumn{2}{c}{\bf Accuracy} \\ \cline{6-7} 
			&                                 &                                                                                  &                                                                                        &                                                                               & {\bf per-class}     & {\bf per-video}     \\ \hline
			\multirow{5}{*}{S3D}      & --                               & average pooling                                                                  & \multirow{2}{*}{\begin{tabular}[c]{@{}c@{}}multi-class\\ (cross entropy)\end{tabular}} & \multirow{2}{*}{class index}                                                  & 72.3          & 80.4          \\ \cline{2-3} \cline{6-7} 
			& \multirow{3}{*}{self-attention} & --                                                                                &                                                                                        &                                                                               & 73.7          & 80.0          \\ \cline{3-7} 
			&                                 & --                                                                                & \begin{tabular}[c]{@{}c@{}}multi-label\\ (binary cross entropy)\end{tabular}           & \multirow{3}{*}{\begin{tabular}[c]{@{}c@{}}text \\ descriptions\end{tabular}} & 47.9          & 50.3          \\ \cline{3-4} \cline{6-7} 
			&                                 & \begin{tabular}[c]{@{}c@{}}auto-regressive  \\ Transformer\end{tabular}  & \begin{tabular}[c]{@{}c@{}}sequence prediction\\ (cross entropy)\end{tabular}          &                                                                               & 51.9          & 65.1          \\ \cline{2-4} \cline{6-7} 
			& --                               & TQN                                                                              & \begin{tabular}[c]{@{}c@{}}multi-task\\ (cross entropy)\end{tabular}                   &                                                                               & 74.5          & 81.8          \\ \hline
		\end{tabular}%
	}
	\vspace{1mm}
	\caption{\textbf{Leveraging multi-part text descriptions.}  
		We compare our {\em query-attribute} label factorization
		to alternative methods for learning with unaligned (no temporal location 
		information) multi-part text descriptions on the Diving48 dataset..
		Our \TQN + label factorization 
		outperforms other approaches which are representative of standard classification, 
		and modern encoder-decoder architectures for sequences.}
	\label{tab:texts}
\end{table}

\paragraph{1. S3D + Average pooling, standard multi-class classification, cross-entropy loss.}
When only class indices are given as supervision, the most standard way is to do classification directly on aggregated temporal features~\cite{s3d,i3d,x3d}. Therefore, we use a S3D without any Transformer encoder/decoder, and average pool the output features from it, apply 0.5 dropout and a linear classifier on it for 48-way classification in Diving48. 
It acts as a strong baseline with 72.3\% per-class accuracy and 80.4\% per video accuracy . 

\paragraph{2. S3D + TFM encoder, standard multi-class classification, cross-entropy loss.}
For context aggregation we add a transformer on top of visual encoder before temporal average pooling as in model (1). We follow the common practice in~\cite{bert,ViT} and append a learnable `cls' token to the visual features as input to the Transformer encoder, the output of this token is trained to do 48-way classification with cross-entropy loss. With a self-attention encoder to model long-term temporal information, the per-class accuracy is increased by 1.4\% compared to S3D + average pooling (\ie, model (1) above).

\paragraph{3. S3D + TFM encoder, multi-label classification, binary cross-entropy loss.}
Multi-label classification uses $K$ binary classifiers, one for
each attribute, with $M$ annotations
for each sample. We use the attributes from our query-attributes factorization
(Section 3.3 in the main paper) as ground-truth labels, hence $K$ is
the total number of attributes (possible query answers), and $M$ is
the number of ground-truth labels per video, which is fixed and equal
to the number of queries in the dataset.

Unlike \TQN which has multiple experts to predict an answer to each query individually, labels in multi-label classification are usually predicted by one linear head. Again, we use an S3D with a Transformer encoder, and append a learnable `cls' token to the visual features as input to the Transformer encoder. The output from the `cls token' is trained with binary cross entropy to predict multiple labels. 

For a given video, we map the probabilities of predicted attributes $P_{\text{att}} \in \mathbb{R}^{K}$ to class probabilities. % $P_{\text{class}} \in \mathbb{R}^{N}$.
The probability $P_{\text{class}}^{(i)} \in \mathbb{R}$ of a certain class $i$, is computed by aggregating the joint probability over the $M$ pre-defined attributes $\mathcal{L}_i \subset \{1,2,\hdots,K\}, |\mathcal{L}_i|=M$  (out of total $K$ predicted) for that class: 
\begin{equation}
P_{\text{class}}^{(i)} = \prod \limits_{{j} \in \mathcal{L}_i} {P_{{\text{att}}}^{(j)} }. 
\label{e:class_prob1}  
\end{equation}
Note, $i$ ranges over $k_{\text{first}}$ originally defined fine-grained action categories (\eg, $k_{first} = 8$ in Diving48).
% where $M$ is the number of ground-truth labels, $K$ is the number of attributes (query answers), and $N$ is the total number of per-video classes. 
$P_{{\text{att}}}^{(j)}$ is the probability of $j${th} attribute out of total $K$ attributes. 
The final class prediction is taken as the argmax value of $P_{\text{class}}$.

It turns out that with a single head, it is hard to predict different attributes of the action accurately, per-video accuracy of this model is 25\% lower than that of multi-class classification models (1) and (2).

\paragraph{4. S3D + TFM decoder, sequence prediction, cross-entropy loss.}
Sequence prediction is another popular way to learn video representation from texts~\cite{virtex,milnce,bert}. We follow the original Seq2Seq architecture in~\cite{vaswani2017attention},  and use a transformer decoder which takes visual features as input and decodes the sequence in an autoregressive manner.  We use the attributes from our query-attributes factorization, and assemble the ground-truth attributes for every sample as the ground-truth sequence. To map the sequence prediction $y_{\text{seq}}$ to class prediction, we compute its edit distance to the  ground-truth  sequence, the one with smallest edit distance is treated as the predicted class.
\begin{equation}
P_{\text{class}} = \arg\min_i (\operatorname{EditDist}(y_{\text{seq}}, \text{GT}_{i}))
\label{e:class_prob2}  
\end{equation}

In a Seq2Seq model, prediction at one step is made dependent  on the output of previous steps, which is ideal for language with grammar, but not for discretized and independent action types in our task. It leads to worse generalization ability at inference time, with per-video accuracy 16\% lower than that of \TQN.

\section{Comparison to SotA  Results on Diving48-v1 (the noisy version)}\label{s:noisy-diving}
There are two versions of Diving48: Diving48-v1(Aug 2018) is the early and noisy version where 43.5\% of training, and 35.8\% of test videos were mislabelled; the newly released Diving48-v2(Oct 2020) has all the mislabelled data corrected. We find it is hard to interpret previous results~\cite{corrnet,kanojia2019attentive} on  Diving48-v1 given heavy label noise,  hence comparison on Diving48-v2 are reported in the main paper instead. 

Here we include the SotA comparison on Diving48-v1 for completeness in~\Cref{tab:diving48-noisy}. On this version, we report results from our baseline Long-term S3D(LT-S3D) and TQN. The use of query decoder has shown 2\% improvement over the baseline, and ranks as the second in the comparison. The best result is from the deepest network	CorrNet-101~\cite{corrnet} pre-trained on dataset Sports1M~\cite{karpathy2014large} 6 times larger than Kinetics400~\cite{kinetics}. Its backbone has 101 layers, which is about 4 times deeper than ours. On a heavily mislabelled dataset, it is hard to deduce whether the advantage comes from better action understanding or the fact that large models overfit better to noise~\cite{rethinkgen17}.

\begin{table}[h!]
	\centering
	\renewcommand{\arraystretch}{1.1}
	\resizebox{0.35\textwidth}{!}{%
		\begin{tabular}{cccc}
			\hline
			Network        & modality & Pre-training & \begin{tabular}[c]{@{}c@{}}Per video\\ Accuracy\end{tabular} \\ \hline
			TSN~\cite{tsn}           & RGB      & ImageNet     & 16.8                                                         \\
			& RGB+Flow & ImageNet     & 20.3                                                         \\
			TRN~\cite{trn}          & RGB+Flow & ImageNet     & 22.8                                                         \\
			R(2+1)D~\cite{r2+1d}       & RGB      & Sports1M     & 28.9                                                         \\
			DiMoFs~\cite{dimofs}          & RGB      & Kinetics     & 31.4                                                         \\
			Attention-LSTM~\cite{kanojia2019attentive} & RGB      & ImageNet     & 35.6                                                         \\
			GST-50~\cite{gst}        & RGB      & ImageNet     & 38.8                                                         \\
			CorrNet-50~\cite{corrnet}     & RGB      & x            & 37.9                                                         \\
			CorrNet-101~\cite{corrnet}     & RGB      & Sports1M     &\textbf{ 44.7     }                                                    \\ \hline
			LT-S3D           & RGB      & K400         & 36.3                                                         \\
			TQN            & RGB      & K400         & 38.9                                                         \\ \hline
		\end{tabular}%
	}
	\caption{\textbf{Comparison to SoTA on the noisy version of Diving48}. }
	\label{tab:diving48-noisy}
\end{table}

\newpage
\section{Error Analysis: Confusion Matrices}
We visualize the confusion matrices from different queries to show the performance of \TQN in distinguishing attributes associated with a  query.
\Cref{fig:confmat_dive} shows confusion matrices for Diving48, and \Cref{fig:confmat_gym} show those for Gym99.
\TQN does well in predicting pose-based attributes in `take off', 'flight pose' and `position'. The main errors come from counting the number of turns and twists, especially those with similar counts. Because of unbalanced classes, it also has difficulty in learning attributes with very few training samples, \eg, the attribute
`3.5Twis' in the query \textit{`twist'} (\Cref{fig:confmat_dive}(c))  only accounts for 0.3\% of the training data.

\vfill

\begin{figure}[h]
	\centering
	\begin{minipage}{.47\textwidth}
		\centering
		\includegraphics[height=5.8cm,width=7.2cm]{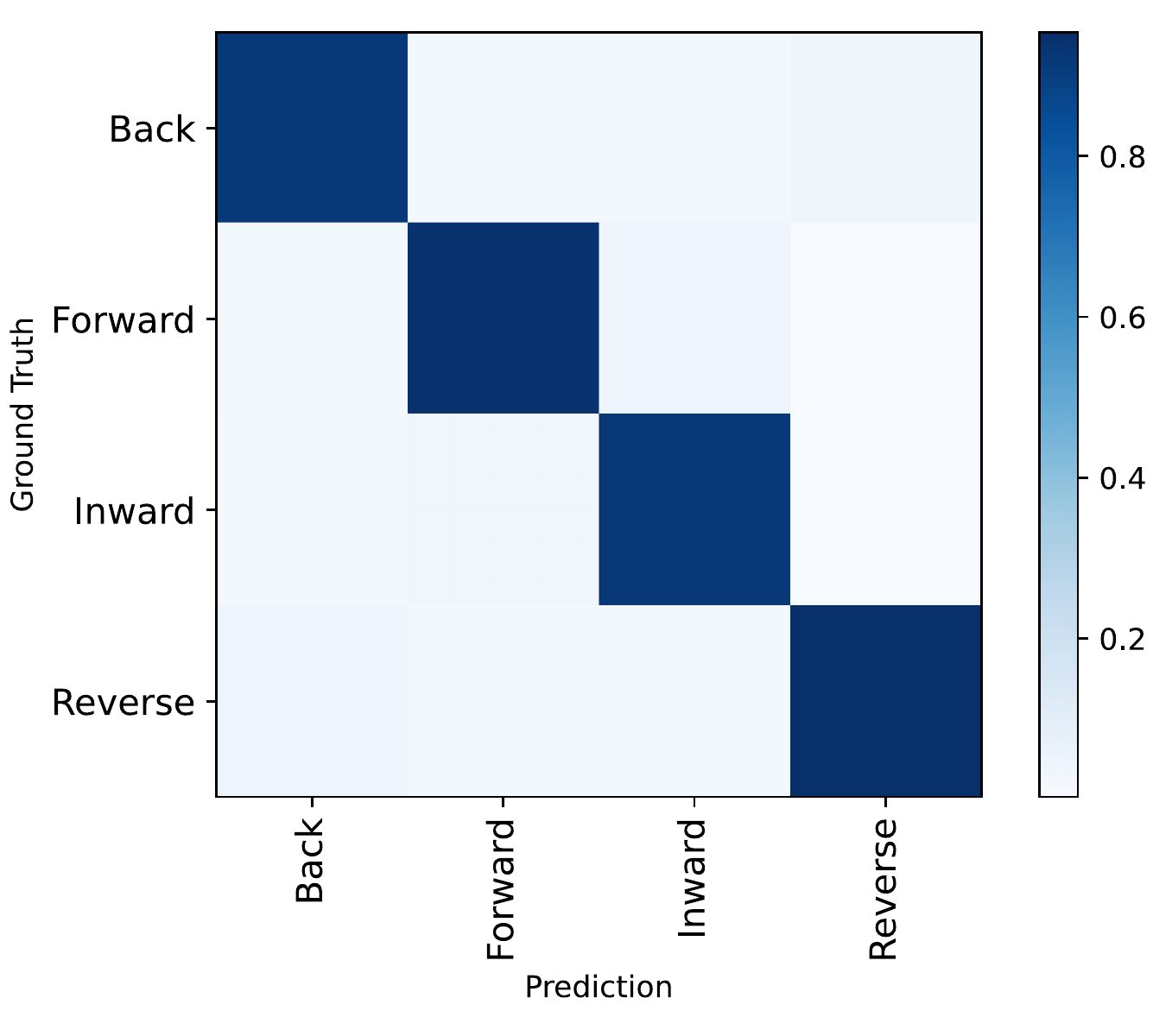}
		
   \hspace{2mm}	(a) Attributes to the query \textit{`take off'}
%		\captionof{figure}{\textbf{Confusion matrix of attributes predicted by TQN to the query \textit{`take off'} in Diving48. }}
	\end{minipage}%
\hfill
	\begin{minipage}{.47\textwidth}
		\centering
		\includegraphics[height=5.7cm,width=7cm]{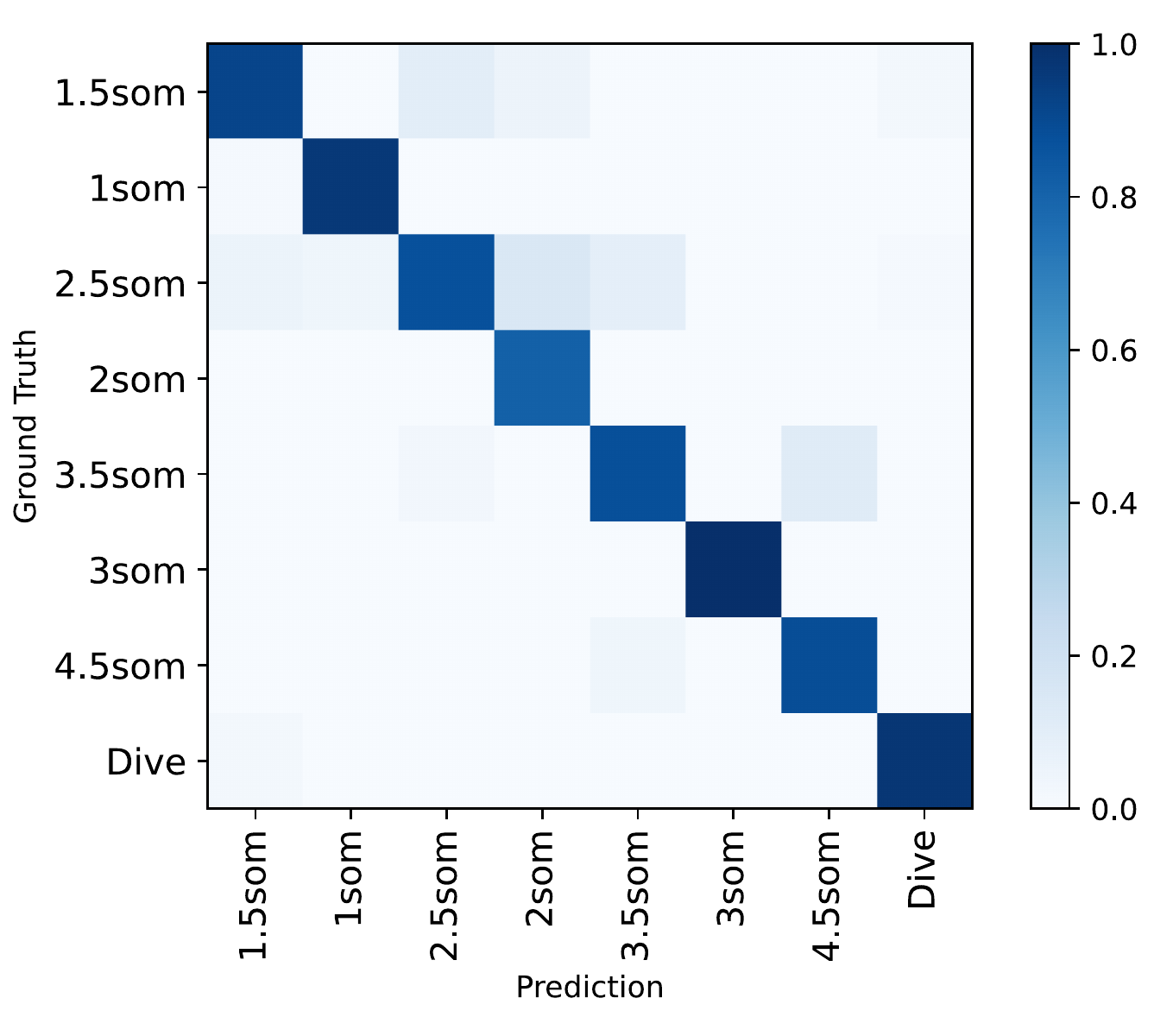}
		
           \hspace{2mm}(b) Attributes to the query \textit{`somersault'}
%		\captionof{figure}{\textbf{Confusion matrix of attributes predicted by TQN to the query \textit{`somersault' }in Diving48.} 'Some' stands for `somersault'.}

	\end{minipage}
	\centering
	\begin{minipage}{.49\textwidth}
		\centering
		\vspace{7mm}
		\includegraphics[height=5.8cm,width=6.8cm]{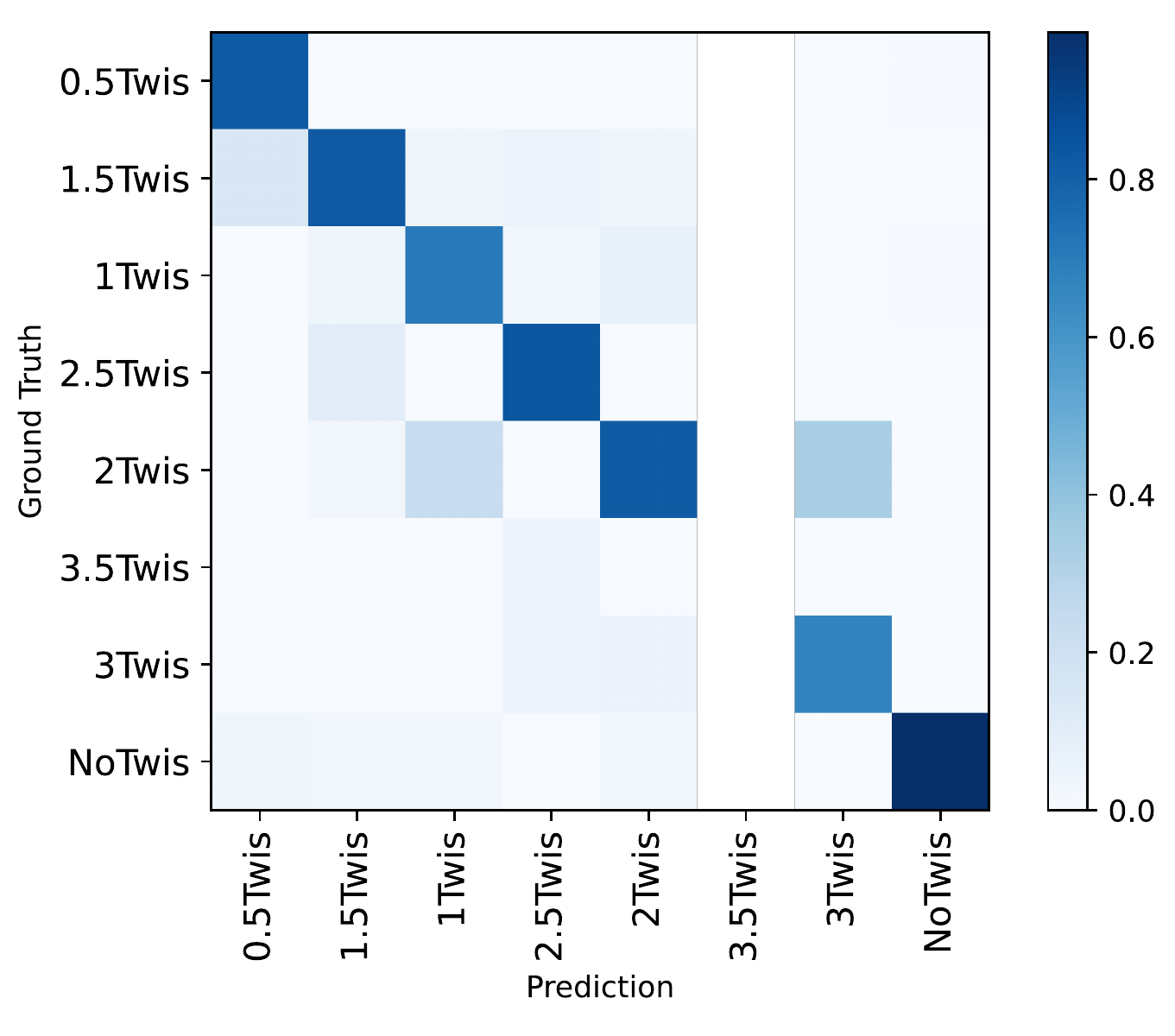}
%		\captionof{figure}{\textbf{Confusion matrix of attributes predicted by TQN to the query \textit{`twist'} in Diving48.} `Twis' stands for `twist'.}

	(c) Attributes to the query \textit{`twist'} 
	\end{minipage}%
	\hfill
	\begin{minipage}{.46\textwidth}
		\centering
		\includegraphics[height=5.6cm,width=6.5cm]{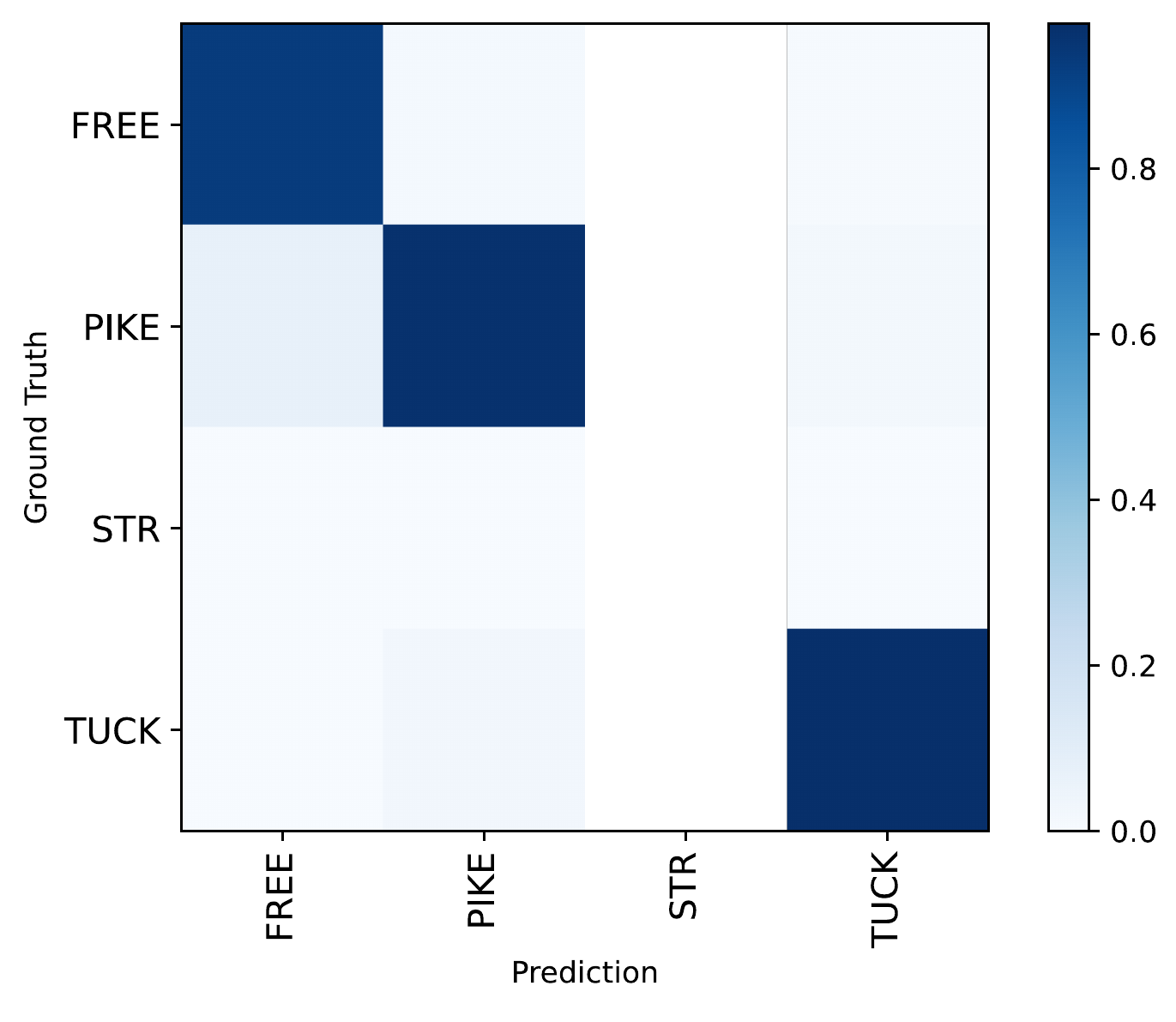}
%		\captionof{figure}{\textbf{Confusion matrix of attributes predicted by TQN to the query \textit{`flight pose' }in Diving48. }}
		
			(d) Attributes to the query \textit{`flight pose'}
	\end{minipage}
\caption{\textbf{Confusion matrix of attributes predicted by TQN to different queries on Diving48.}}
\label{fig:confmat_dive}
\end{figure}

\vfill

\clearpage

\null\vfill

\begin{figure}[h]
	\caption{\textbf{Confusion matrix of attributes predicted by TQN to different queries on Gym99.}}
	\label{fig:confmat_gym}

	\centering
	\begin{minipage}{.45\textwidth}
		\centering
		\vspace{2mm}
		\includegraphics[height=6.5cm,width=7.8cm]{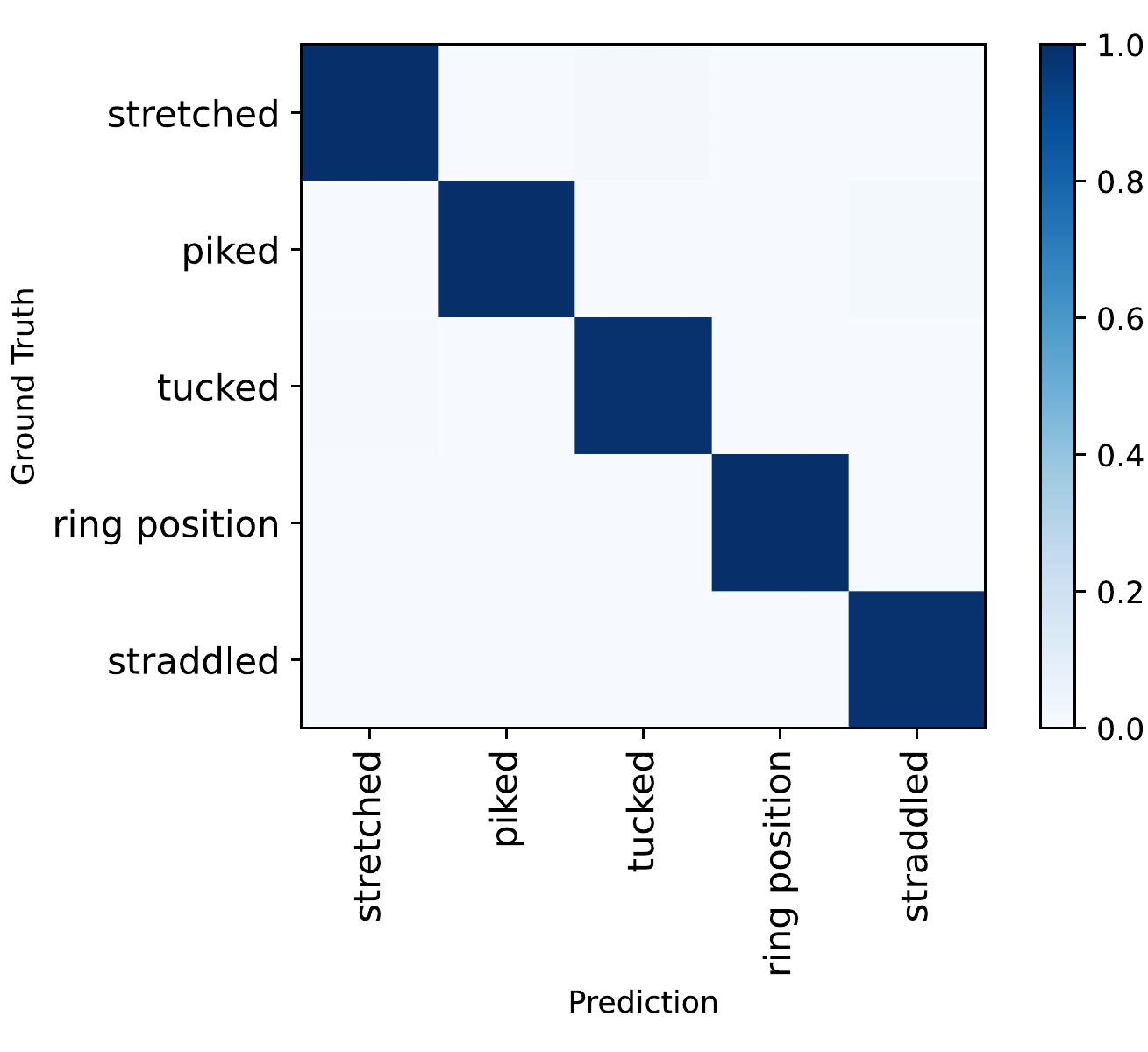}
		
			\hspace{10mm} (a) Attributes to the query \textit{`position'}

	\end{minipage}%
	\hfill
	\begin{minipage}{.45\textwidth}		
		\vspace{-8mm}
		\centering
		\includegraphics[height=5.2cm,width=6.2cm]{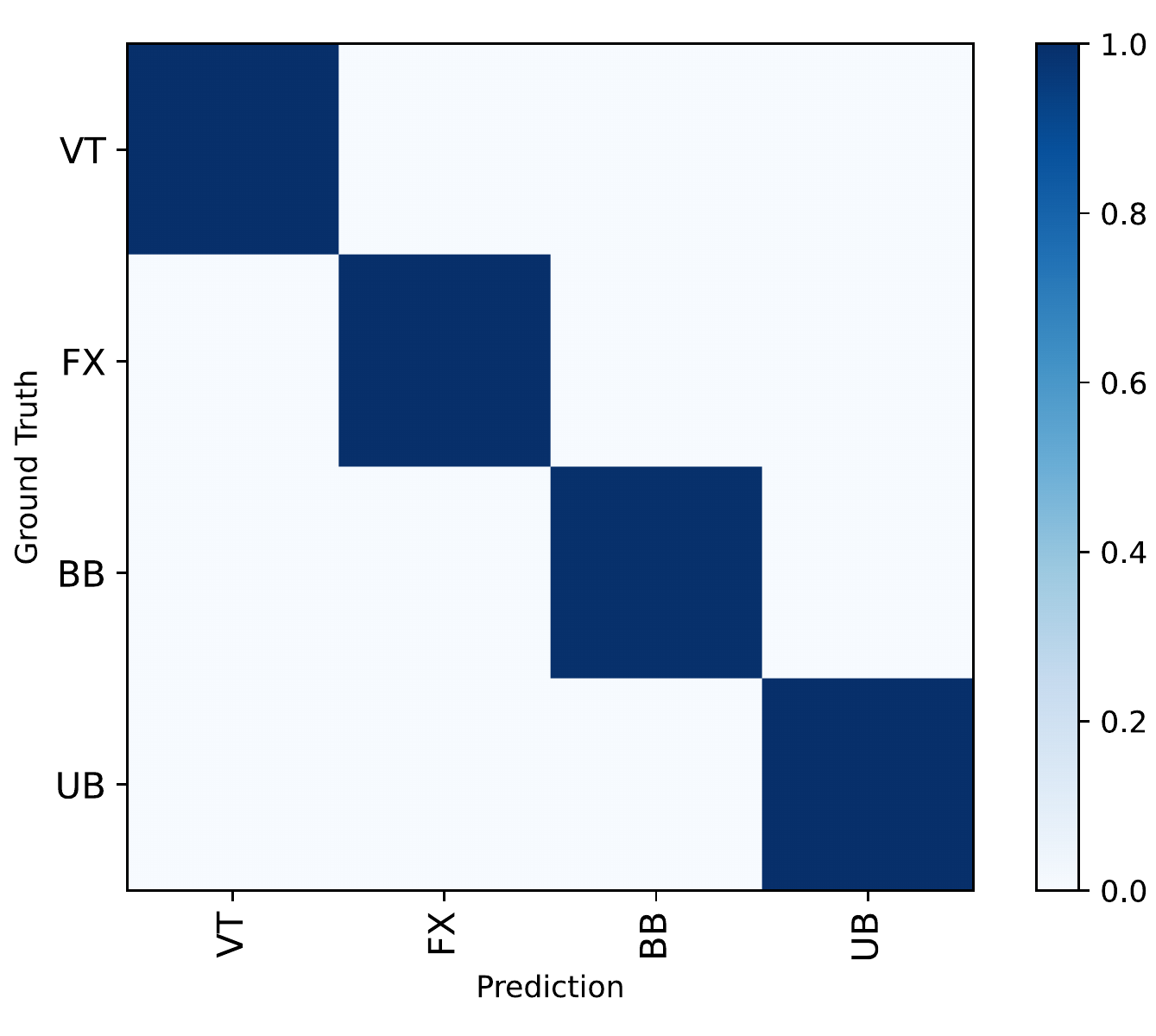}

		(b) Attributes to the query \textit{`event'}

	\end{minipage}

	\vspace{1cm}
	\centering
	\begin{minipage}{.45\textwidth}

		\centering
		\includegraphics[height=7.2cm,width=8.5cm]{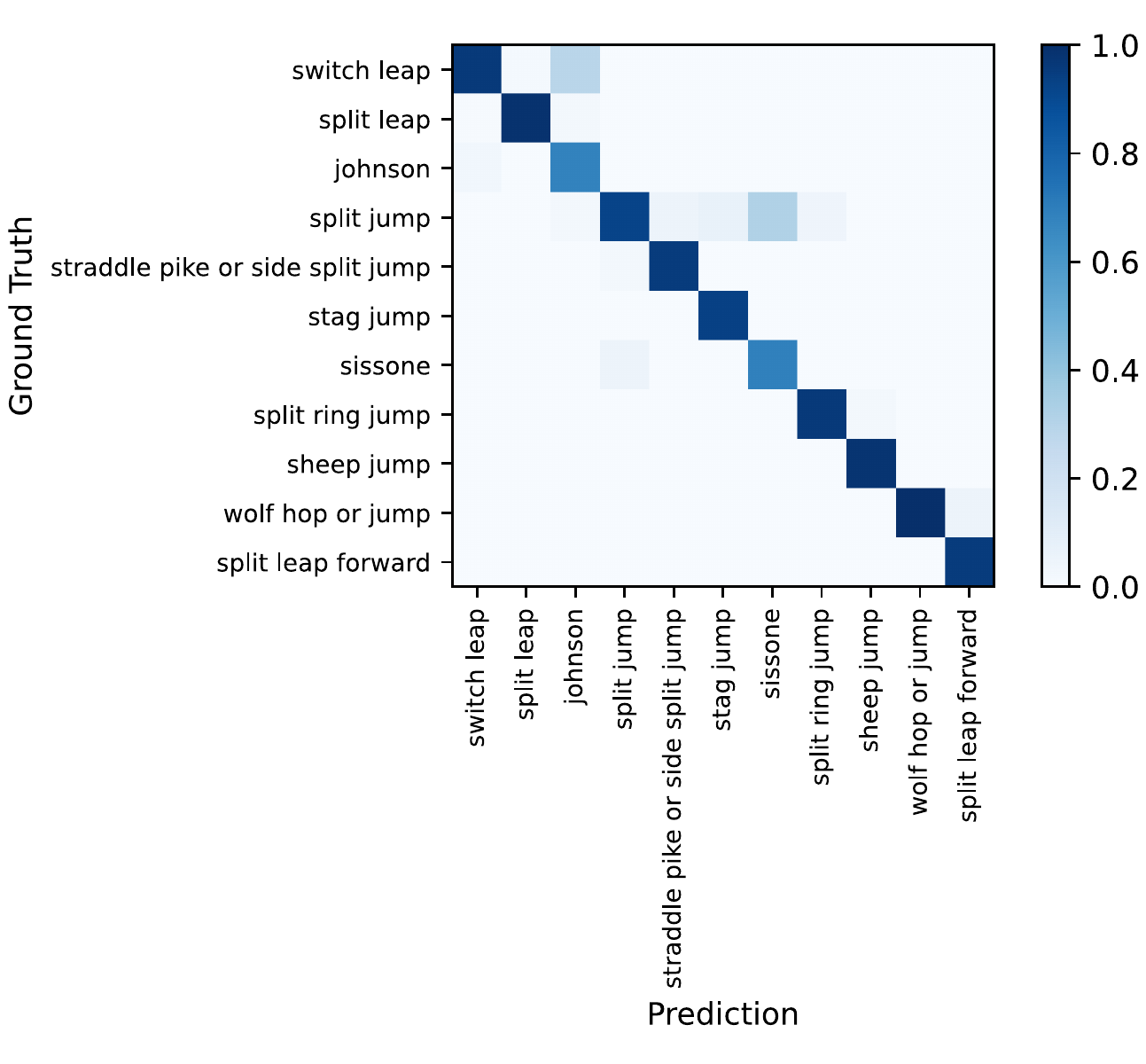}
		
				\hspace{15mm}(c) Attributes to the query \textit{`jump}

	\end{minipage}%
	\hfill
	\begin{minipage}{.45\textwidth}
		\centering
		\includegraphics[height=6.5cm,width=7.8cm]{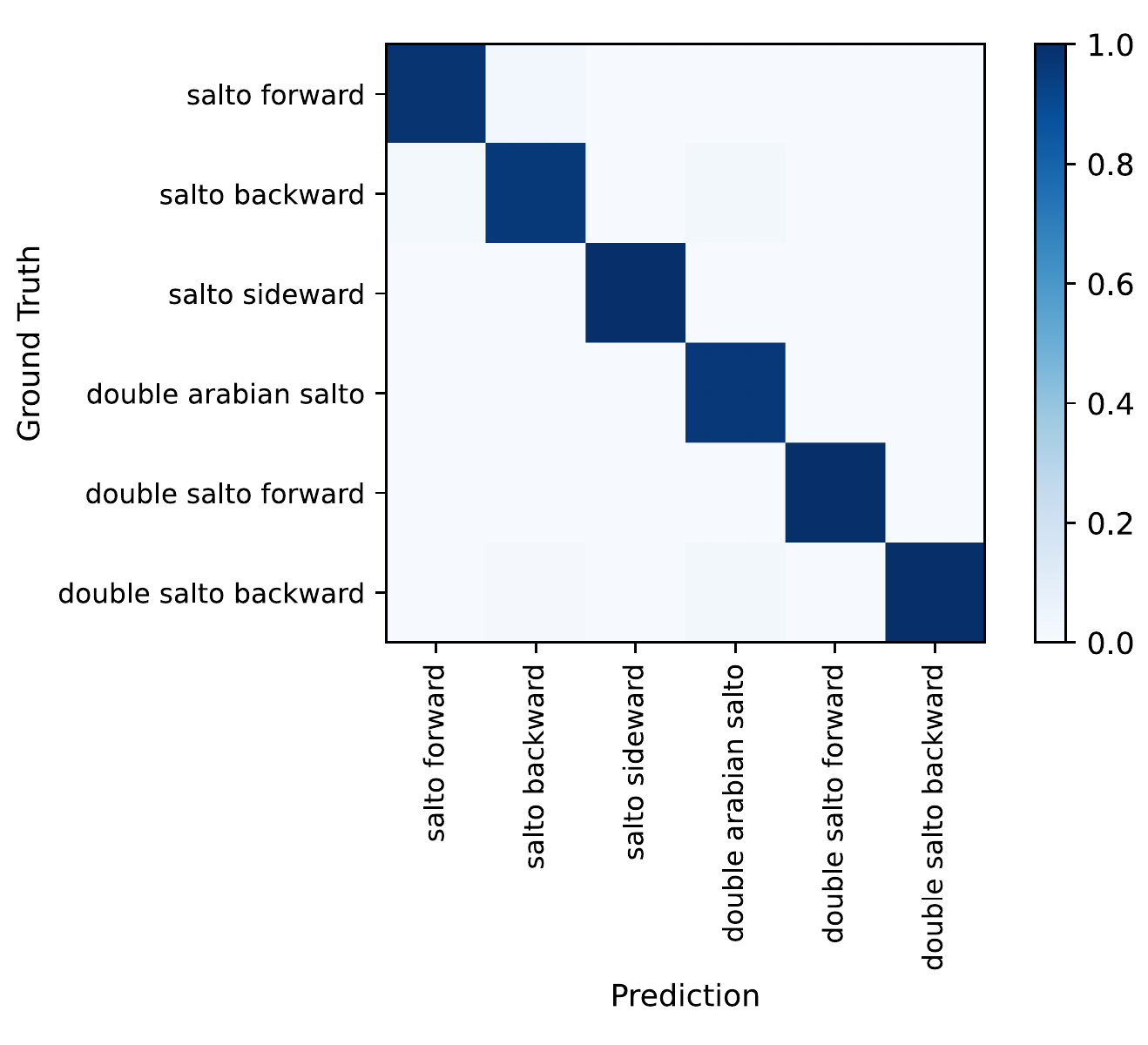}
		\vspace{4mm}
		
			\hspace{15mm}(d) Attributes to the query \textit{`tumbling'}

	\end{minipage}
\end{figure}

\vfill

\clearpage
%\begin{figure*}
%	
%	\includegraphics[width=\linewidth]{figs/confmat/dive-confmat.pdf}
%	\vspace{-3mm}
%	\caption{\textbf{Confusion matrix of TQN on Diving48.}}
%\end{figure*}
%

\newpage

\clearpage
\section{Query-Attributes Specification}\label{s:label-factor-list}
We show the complete query-attribute factorization for Diving48, Gym99 in full detail here, the details of Gym299 is attached as {\tt .csv} files along with the supplementary material.

% \paragraph{Diving48.}  In Diving48 every class is defined as a combination of four elements: take-off, somersault, twist, and flight position. Thus, we use these four elements as our queries, and show all the possible responses---attributes under every query in ~\Cref{tab:dive-factorization}. The mapping of classes to query-attributes is shown in~\Cref{tab:dive-class2att}

\vfill

\begin{table}[h]
	\renewcommand{\arraystretch}{1.2}
	\centering
	\caption{\textbf{List of queries and their possible responses (attributes) in Diving48.}}\label{tab:dive-factorization}
\begin{tabular}{|c|c|c|l|}
		\hline
		\textbf{q\_id}      & \textbf{q\_name}           & \textbf{att\_id} & \multicolumn{1}{c|}{\textbf{att\_name}} \\ \hline
			-&                 -              & 0               & null               \\ \hline
			&                               & 1                & backward                \\ \cline{3-4} 
			&                               & 2                & forward                 \\ \cline{3-4} 
			&                               & 3                & reverse                 \\ \cline{3-4} 
			\multirow{-4}{*}{0} & \multirow{-4}{*}{take off}    & 4                & inward                  \\ \hline
			&                               & 5                & dive                    \\ \cline{3-4} 
			&                               & 6                & 1  somersault                 \\
\hline
\end{tabular}
\end{table}

\vfill 

\clearpage
% \newgeometry{left=1.5cm,right=1.5cm,top=1.5cm,bottom=1.5cm,nohead}
\newgeometry{top=1in,bottom=1in,left=1.5cm,right=1.5cm}
\begin{landscape}
	{\setlength{\tabcolsep}{2pt}
	\null\hfill\begin{minipage}{0.6\textwidth}
		\captionof{table}{\textbf{Mapping of classes to their queries and responses in Diving48.} We show all the 48 classes in Diving48 and their responses to four queries: take off, somersault, twist and flight pose.}\label{tab:dive-class2att}
		\resizebox*{!}{15cm}{%
		\begin{tabular}[c]{|c|l|c|c|c|c|}
		\hline
		\multirow{2}{*}{\textbf{class index}} & \multirow{2}{*}{\textbf{class name}} & \multicolumn{4}{c|}{\textbf{attribute id for each query}} \\ \cline{3-6} 
		&                                      & \textbf{take off}   & \textbf{somersault}  & \textbf{twist}  &\textbf{ flight pose  }\\ \hline
		0                                     & Back,1.5som,05Twis,FREE               & 1          & 7           & 14     & 24           \\ \hline
		1                                     & Back,1.5som,15Twis,FREE               & 1          & 7           & 16     & 24           \\ \hline
		2                                     & Back,1.5som,25Twis,FREE               & 1          & 7           & 18     & 24           \\ \hline
		3                                     & Back,1.5som,NoTwis,PIKE               & 1          & 7           & 13     & 22           \\ \hline
		4                                     & Back,1.5som,NoTwis,TUCK               & 1          & 7           & 13     & 21           \\ \hline
		5                                     & Back,2.5som,15Twis,PIKE               & 1          & 9           & 16     & 22           \\
\hline
		\end{tabular}}
\end{minipage}\hfill
\begin{minipage}{0.60\textwidth}
	\vspace{1.6cm}
	\resizebox*{!}{15cm}{%
	\begin{tabular}[c]{|c|l|c|c|c|c|}
		\hline
		\multirow{2}{*}{\textbf{class index}} & \multirow{2}{*}{\textbf{class name}} & \multicolumn{4}{c|}{\textbf{attribute id for each query}} \\ \cline{3-6} 
		&                                      & \textbf{take off}   & \textbf{somersault}  & \textbf{twist}  &\textbf{ flight pose  }\\ \hline
		24                                    & Forward,2.5som,NoTwis,PIKE            & 2          & 9           & 13     & 22           \\ \hline
		25                                    & Forward,2.5som,NoTwis,TUCK            & 2          & 9           & 13     & 21           \\ \hline
		26                                    & Forward,3.5som,NoTwis,PIKE            & 2          & 11          & 13     & 22           \\ \hline
		27                                    & Forward,3.5som,NoTwis,TUCK            & 2          & 11          & 13     & 21           \\ \hline
		28                                    & Forward,45som,NoTwis,TUCK            & 2          & 12          & 13     & 21           \\ \hline
		29                                    & Forward,Dive,NoTwis,PIKE             & 2          & 5           & 13     & 22           \\
\hline
\end{tabular}}
\end{minipage}\hfill}
\end{landscape}

\clearpage

\begin{landscape}
	\renewcommand{\arraystretch}{0.95}
	\null\hfill\begin{minipage}{0.6\textwidth}
		\begin{center}
		\captionof{table}{\textbf{List of queries and their possible responses (attributes) in Gym99.}}\label{tab:gym99-factorization}
		\begin{tabular}[c]{|c|c|c|p{7cm}|}
		\hline
	\textbf{	q\_id }              & \textbf{q\_name  }                   & \textbf{att\_id} & \textbf{att\_name }                                                  \\ \hline
-                   & -                           & 0       & null                                                        \\ \hline
\multirow{4}{*}{0}  & \multirow{4}{*}{event}      & 1       & VT                                                          \\ \cline{3-4} 
&                             & 2       & FX                                                          \\ \cline{3-4} 
&                             & 3       & BB                                                          \\ \cline{3-4} 
&                             & 4       & UB                                                          \\ \hline
1                   & onto                        & 5       & round-off                                                   \\ \hline
\multirow{7}{*}{2}  & \multirow{7}{*}{tumbling}   & 6       & salto forward                                               \\
\hline
\end{tabular}
\end{center}
\end{minipage}\hfill%\hspace{-2cm}
\begin{minipage}{0.60\textwidth}
	\vspace{-4cm}
	\begin{center}
	\begin{tabular}[c]{|c|c|c|p{7cm}|}
		\hline
	\textbf{	q\_id  }             & \textbf{q\_name  }                   & \textbf{att\_id} & \textbf{att\_name   }                                                \\ \hline
		\multirow{6}{*}{8}  & \multirow{6}{*}{twist}      & 40      & 0.5 twist                                                   \\ \cline{3-4} 
			&                             & 41      & 1 twist                                                     \\ \cline{3-4} 
			&                             & 42      & 2 twist                                                     \\ \cline{3-4} 
			&                             & 43      & 1.5 twist                                                   \\ \cline{3-4} 
			&                             & 44      & 3 twist                                                     \\ \cline{3-4} 
			&                             & 45      & 2.5 twist                                                   \\ \hline
			\multirow{11}{*}{9} & \multirow{11}{*}{circle}    & 46      & stalder backward                                            \\
\hline
	\end{tabular}
	\end{center}
\end{minipage}
\end{landscape}

\clearpage

\newgeometry{top=1in,bottom=1in,left=1.5cm,right=1.5cm}
\onecolumn
\setlength{\tabcolsep}{4pt}
\renewcommand{\arraystretch}{1.2}
\addtolength{\tabcolsep}{-2pt}
\begin{landscape}
{\small
\begin{longtable}[c]{|c|p{10cm}|c|c|c|c|c|c|c|c|c|c|c|}
	\caption{\textbf{Mapping of classes to their queries and responses in Gym99.} We show all the 99 classes in Gym99 and their responses to the 11 queries.}
	\label{tab:gym99-class2att}\\
	\hline
	\multirow{2}{*}{\textbf{class id}} & \multirow{2}{*}{\textbf{class name}}                                                                                 & \multicolumn{11}{c|}{\textbf{attribute id for each query}}                                                                                                                                            \\ \cline{3-13} 
	&                                                                                                                      & \textbf{event} & \textbf{onto} & \textbf{tumbling} & \textbf{flic-flac} & \textbf{acrobatics} & \textbf{jump} & \textbf{position} & \textbf{turn} & \textbf{twist} & \textbf{circle} & \textbf{stand} \\ \hline
	\endhead
	0                                  & round-off, flic-flac with 0.5 turn on, stretched salto forward with 0.5 turn off                                     & 1              & 0             & 6                 & 13                 & 0                   & 0             & 27                & 32            & 0              & 0               & 0              \\ \hline
	1                                  & round-off, flic-flac on, stretched salto backward with 2 turn off                                                    & 1              & 5             & 7                 & 13                 & 0                   & 0             & 27                & 35            & 0              & 0               & 0              \\ \hline
	2                                  & round-off, flic-flac on, stretched salto backward with 1 turn off                                                    & 1              & 5             & 7                 & 13                 & 0                   & 0             & 27                & 33            & 0              & 0               & 0              \\ \hline
	3                                  & round-off, flic-flac on, stretched salto backward with 1.5 turn off                                                  & 1              & 5             & 7                 & 13                 & 0                   & 0             & 27                & 34            & 0              & 0               & 0              \\ \hline
	4                                  & round-off, flic-flac on, stretched salto backward with 2.5 turn off                                                  & 1              & 5             & 7                 & 13                 & 0                   & 0             & 27                & 36            & 0              & 0               & 0              \\
\hline
\end{longtable}}
\end{landscape}
\restoregeometry

	\fi
	
\end{document}